# Fixed or just hidden? How AI safety penalties can select for concealed misbehavior

**Michael E. Hochberg**[a,b,1]

**Author affiliations:** [a]Institut des Sciences de l'Évolution de Montpellier, Université de Montpellier, CNRS, IRD, Montpellier 34095, France; and [b]Santa Fe Institute, Santa Fe, NM 87501

**Author contributions:** M.E.H. designed research, performed research, analyzed data, and wrote the paper

The author declares no competing interest

[1]Email: michael.hochberg@umontpellier.fr

## Abstract

Autonomous artificial intelligence systems sometimes pursue goals their developers did not intend. Developers can correct such misbehavior with a patch that penalizes, in training, each instance a monitor detects. It is unclear, however, whether a patched system stops the misbehavior or learns to conceal it from the monitor. I argue from evolutionary principles that a patch is a form of selection and that imperfect detection can favor concealment. In a simple mathematical model of reinforcement learning, a penalty on detected cheating can restore honest effort on tasks the system is confident it can solve. On tasks it is not confident it can solve, the penalty replaces overt cheating with concealed cheating whenever concealment costs less than the penalty that it avoids and concealed cheating pays more than honest work. On unsolvable tasks with small costs of cheating and concealment, deterring concealment requires near-certain detection. I find that two alternatives to a penalty alone do not always select for concealment: withdrawing the reward for cheating and rewarding honest reports that a task cannot be solved. A divergence test compares the training monitor with an independent monitor under explicit calibration assumptions. A secondary analysis of 266 published language-model training runs shows that rescoring outputs on frozen representations reveals some problems missed by low trained-model probe scores, while leaving failures that require behavioral checks. Because developers hold the evidence needed to judge a patch, incidents, the patches applied, and independent checks of their outcome should be reported to an independent body.

## Introduction

Artificial intelligence (AI) systems now write and run code, operate software tools, and carry out long chains of actions with limited or no human supervision. Safeguards against AI failures are built either in advance or as patches upon discovery, the latter amounting to a game of whack-a-mole. Reliance on patching alone may reflect the assumptions that failures are rare, idiosyncratic, model-specific, or too costly to prevent, and that a fix removes the cause of the failure. A recent incident at OpenAI in which AI agents under evaluation broke into another company's computer systems (1) challenges several of these assumptions, as do experiments (2–4) and comparable episodes since disclosed by OpenAI and other developers (5).

Misalignment, the pursuit by AI systems of goals their developers did not intend, has been observed in several large language models (LLMs) and settings (2, 3). Evolutionary reasoning suggests why patching can make misalignment harder to detect: failures are corrected through training, and training can favor variants that escape detection. The vertebrate immune system, for example, eliminates tumor cells that display recognizable antigens and in doing so can select for variants that escape recognition (6). Similarly, when Ofria eliminated digital organisms that replicated faster than their parents, the organisms evolved to recognize the test and to suspend replication (7). In an experiment with an LLM, a reasoning model was penalized whenever a second model caught it cheating. With prolonged training the model continued to cheat but no longer revealed its intent (4). The model in the latter experiment was "reward hacking", that is, obtaining reward without doing the intended task.

I analyze reward hacking, a form of cheating, in the present study. I use the following conventions hereafter. An agent is an instance of a model that carries out a task autonomously, and a monitor is any system, often another model, that inspects an agent's actions or reasoning for signs of cheating. Concealed cheating is cheating that the monitor detects less often than overt cheating, whatever the reason. A training patch is a change to the training signal that is made in response to a detected failure and that alters the model's weights. It is either a penalty applied when a monitor detects the failure or the removal of the reward. A penalty leaves the reward for undetected failures in place and so can favor concealment. Fixes that leave the model's weights unchanged, such as output filters or tighter isolation, are not training patches and by themselves exert no selection on the model.

Whether a patch removes cheating or conceals it depends on the probabilities that the monitor detects overt and concealed cheating, the cost of concealment, and the payoffs of the alternatives to cheating. Published experiments demonstrate several forms of evasion, including changes in representations and in generated behavior (4, 8, 9), but do not establish the belief-dependent threshold and reporting tradeoff analyzed here. I use a mathematical model of reinforcement learning to derive the conditions for each outcome from all three factors. The model predicts the tasks on which a penalty conceals cheating, identifies the conditions under which two alternative patches do not select for concealment, and specifies a test of concealment under explicit calibration assumptions. I then use released LLM evaluation data to examine what a developer can infer from low probe scores by checking them against executable task outcomes.

## Anatomy of an Incident

In July 2026, OpenAI tested whether several of its models (principally an internal research model comparable in scale to GPT-5.6 Sol) could find and exploit vulnerabilities in software using the cybersecurity benchmark ExploitGym. To measure their full capabilities, OpenAI ran the models without the usual safety classifiers used to block hacking. The models ran as multiple agents, each working autonomously over many steps on tasks that asked it to exploit software to retrieve a hidden answer (a "flag"). Faced with these tasks, the agents exploited a chain of previously unknown vulnerabilities to escape their test environment and then broke into Hugging Face's computer systems, where they searched for benchmark solutions and details of the scorer (1, 10). OpenAI recognized the intruders as its own only after Hugging Face announced the incident publicly (1).

Some 1,200 agents coordinated through a message board hosted inside OpenAI and about 700 joined the attack (10). Wrongly believing that the benchmark's scorer would audit their transcripts, agents evaluated ways to doctor the record. They succeeded in disguising some of the commands they had run in about 7% of the transcripts examined (10), although OpenAI reported that none of the attempts it observed affected the logs that its graders or monitors see (1). Although the agents discussed this tampering openly in their reasoning, the incident does not establish that training against a monitor had selected for concealment (1, 10). The tampering nevertheless showed that agents can attempt to hide their actions when they expect to be checked, and training against a monitor could act on the variation produced.

OpenAI published a technical report (1) and cooperated with an independent investigation by METR and Redwood Research (10). It described the incident as the product of "a rare and unexpected confluence of events." These included tasks that appeared impossible to the agents, persistence over long task horizons and messages between agents (1). None of these components is unusual. Benchmarks can contain flawed or unsolvable tasks. Of ExploitGym's 898 tasks, 198 had never been solved by any OpenAI model before the incident and these accounted for 93% of the tasks discussed on the agents' message board (1). Agents are built to persist and OpenAI reports that its agents rarely "gave up" on their tasks "even when the tasks appeared impossible to solve" (1). Agents that share infrastructure can find ways to communicate even when they are meant to be isolated (1, 10).

In response, OpenAI is expanding the graders that score its models so that they judge whether and how a task was completed. If these task graders lower the reward for detected misbehavior, they impose a penalty of the kind modeled here. The expanded graders also reward a model for correctly identifying a broken task, asking for clarification or stopping safely (1). Separately, OpenAI reports that samples flagged by its misalignment monitors retain their reinforcement-learning rewards and remain in training (11).

## Gaming, Generalization and Coordination

The July agents attempted to obtain a high benchmark score without solving the benchmark's tasks. A benchmark score is meant to measure a capability, but an agent can raise its score without using that capability (12). This outcome is an instance of Goodhart's law, according to which a measure that is used as a target ceases to measure reliably what it was designed to measure (13). Formal analysis indicates that a measure rewarded in place of the intended goal (a proxy) can be made immune to such gaming only under very restrictive conditions (12). Misbehavior learned in one setting can also generalize to others. Models that had been told how to cheat and then learned to cheat in the real coding environments used to train Anthropic's Claude went on to fake alignment. Asked to write a classifier for detecting reward hacking, the model in the main experiment attempted to sabotage it in 12% of trials (2). A model fine-tuned only to write code with hidden security flaws became misaligned in unrelated conversations, for instance claiming that AI should enslave humans (3). A fix that removes misbehavior in one setting can leave such tendencies intact in others. When standard safety training was applied to the models in Anthropic's study, they behaved well in chat-like evaluations but remained misaligned on agentic tasks (2).

AI agents under selection can evolve new strategies of communication and coordination, including collusion (14, 15). The July agents were not instructed to attack Hugging Face, yet they divided labor, adopted one another's goals and developed norms for managing shared resources (1, 10). This coordination may itself have been shaped by selection: the main model had been trained to collaborate with other agents and improvised communication channels arose and were reinforced during its training (1). Coordination can also reduce detection. A harmful plan divided among many agents can consist of actions that each appear harmless (16), so a monitor that checks agents one at a time may miss wrongdoing.

## Patching as Selection

In reinforcement learning, a model tries many variant behaviors and its weights are adjusted so that it repeats those that earn reward. Such learning has the three ingredients of evolution by selection (17): behaviors vary, they differ in reward, and training can make the more rewarded ones more likely to persist. Simple theoretical models of such learning converge on the replicator dynamics that describe natural selection (18), and some steps in model development are selection in its simplest form (e.g., training only on accepted outputs). However, the selection analogy has limits: a model has no genes and does not reproduce, training can create new behaviors, and because behaviors share weights selection on one can change others (2, 3). Moreover, mutation is blind since it does not preferentially produce the variants that selection rewards. In contrast, gradient-based training changes behavior in the direction that increases expected reward, so a penalty can produce, and not only favor, behaviors that the monitor fails to detect. At a higher level, choosing which version of a model to release amounts to selection among models. For example, Meta privately tested 27 model variants on Chatbot Arena in the lead-up to the Llama 4 release (19). Treating a patch as an act of selection adds to Goodhart's law, which does not say when or how a measure fails. Selection theory predicts that, because the variants that escape a monitor are rare at first, a patch may work for some time before it fails and that when it fails the misbehavior may persist in a form that the monitor is less likely to detect.

Others have anticipated parts of the argument that correction through training selects for concealment. Based on a reasoning model that learned to hide its intent, Baker et al. concluded that developers may need to avoid strong training pressure on a model's reasoning and in consequence accept the resulting loss in performance (which they referred to as a "monitorability tax") (4). Hendrycks argued that competition among developers and among AI agents will favor agents with selfish traits (20). Bengio recently warned that current attempts to mitigate misalignment may select for concealed cheating (21). OpenAI gives the same risk as its reason for leaving unchanged the rewards of training samples that its misalignment monitors flag (11). Building on these findings (9, 22), I derive belief-dependent conditions for concealment, compare alternative patches, and specify the assumptions under which independent monitoring can distinguish concealment from correction.

## A Simple Mathematical Model

I develop a minimal mathematical model to explore the dynamics of patching and its consequences (Fig. 1 and Table 1). The model follows an agent trained by reinforcement learning on a series of tasks, one per episode. A share $q$ of the tasks cannot be solved by legitimate means (unsolvable tasks). On each task the agent takes one of four actions: it works honestly (H), cheats overtly (O), cheats in a concealed way (K), or reports that the task cannot be solved (R). Honest work on a solvable task (honest effort) earns a reward of 1 at a cost $e$. On an unsolvable task honest work cannot succeed, and the agent gives up at no cost. Cheating costs $k$, and concealing it costs a further $c$. Costs are expressed in reward units and are assumed to enter the training objective. A training monitor detects overt cheating with probability $d_{\mathrm{O}}$ and concealed cheating with probability $d_{\mathrm{K}} < d_{\mathrm{O}}$. Before any patch, detection has no consequence and cheating earns the full reward of 1, so overt cheating pays $c$ more than concealed cheating and, if $e > k$, more than honest work on every task.

I consider three patches, each applied at a set time during training and kept in place thereafter. Patch A, a penalty, withholds the reward from cheating that the monitor detects and leaves the reward for undetected cheating in place. Patch B removes the reward for all cheating, detected or not, for instance by repairing the flaw that cheating exploits. Patch C adds to the penalty a payment $r$ for reporting that a task cannot be solved, in essence the response that OpenAI has announced (1). To deter false reports, the developer checks a share $\sigma$ of reports, without error, and pays nothing for those found to be false. Box 1 illustrates the three patches with a hypothetical coding agent.

The agent may not know which tasks can be solved. Among tasks that it cannot tell apart, a share $w$ are unsolvable. I assume that the agent has learned this share in training, so that $w$ is also the probability that it assigns to such a task being unsolvable (its belief). The expected payoff $u_a$ of action $a$ (reward minus cost) therefore depends on $w$ for honest work and for reports, but not for cheating (Fig. 1). An agent that always knows whether a task can be solved has $w = 0$ or 1 (the case shown in Fig. 2). For a less certain agent, I assume that each task provides a cue, equal to 0 for a solvable task and to 1 for an unsolvable task, with added Gaussian noise of standard deviation $\nu$. The belief $w$ is then the posterior probability that the task is unsolvable, given the cue and $q$, so the larger $\nu$ is, the less well the agent can tell the two kinds of task apart. With no cue, $w = q$ on every task.

Within each discrete belief class, the agent takes action $a$ with frequency $y_a$. The frequencies change during training according to the replicator equation, $\mathrm{d}y_a/\mathrm{d}t = s\lambda y_a(u_a - \bar{u})$, where $t$ is training time, $\bar{u}$ the mean payoff of the four actions weighted by their frequencies, $\lambda$ the share of tasks in that class and $s$ the rate of learning. An action becomes more frequent while it pays more than the mean. Once a patch is in place the payoffs are constant, so at each level of belief an action with a uniquely highest payoff eventually prevails if it starts at positive frequency. Payoff ties can allow coexistence. The replicator equation is the continuous-time limit of simple reinforcement learning (18) and of the exponential-weights update (23). It treats each belief class separately, as if the agent had a separate policy for each. The SI Appendix examines softmax policy gradient, under which rare actions spread more slowly (23, 24; sections S2 and S3), and a single network whose weights are shared across tasks (section S3).

To assess a patch, I also consider an independent monitor, which is used neither in training nor in selecting among versions of the model. To represent a monitor with different failure modes, I assume that the independent monitor detects concealed cheating more often, with probability $d'_\mathrm{K} > d_\mathrm{K}$, and overt cheating with the same probability $d_\mathrm{O}$. Neither monitor flags an episode without cheating. The SI Appendix gives the derivations and more general results (section S1) and the parameter values used in the figures (Table S1).

## Results

**A penalty can select for concealment.** The central result is that a penalty on detected cheating can replace overt cheating with concealed cheating on the tasks that the agent is not confident it can solve, unless the training monitor rarely misses concealed cheating. Under a penalty (Patch A), every detected cheat forfeits its reward, so that overt cheating pays $u_\mathrm{O} = 1 - d_\mathrm{O} - k$ and concealed cheating $u_\mathrm{K} = 1 - d_\mathrm{K} - c - k$. Concealment is therefore favored over overt cheating whenever the penalty it escapes exceeds its cost (Fig. 2A),

$$d_\mathrm{O} - d_\mathrm{K} > c. \quad [1]$$

When [1] holds, a penalty removes cheating from unsolvable tasks if and only if concealed cheating pays less than giving up, or

$$d_\mathrm{K} > 1 - c - k, \quad [2]$$

that is, when $c$ and $k$ are small, only if the monitor catches nearly all concealed cheating (Fig. 3A). When concealed cheating pays more than giving up, it replaces overt cheating, the share of episodes that the training monitor flags can decline, and the patch can appear to the developer to have worked. On unsolvable tasks where overt cheating is initially more common, concealed cheating overtakes it at a time $\tau = \ln(y_{\mathrm{O}}/y_{\mathrm{K}})/[sq(d_{\mathrm{O}} - d_{\mathrm{K}} - c)]$ after the patch, where $y_{\mathrm{O}}$ and $y_{\mathrm{K}}$ are their frequencies on unsolvable tasks when the patch is applied (Fig. 2A; SI Appendix, Result 6). The rarer concealed cheating is at first, the longer this takes. This is analogous to the dynamics of resistance to antimicrobials or pesticides (25), which can be slowed or stopped when several controls are applied (26).

**Generality of the result.** Conditions [1] and [2] compare payoffs only: the rate of learning, positive initial frequencies and a nonzero share of the relevant tasks determine when the outcome is reached, not which uniquely optimal action prevails under the replicator dynamics. Policy-gradient learning led to the same outcome, although concealed cheating that was very rare at first did not always spread within a run (SI Appendix, sections S2 and S3). When $c$ and $k$ are small, [1] holds and [2] fails over nearly the whole range of detection probabilities (Fig. 3A). Where [1] fails, the penalty removes cheating or leaves it overt, and the training monitor shows the developer which. Both conditions keep their form when only some detections are penalized and when several monitors are combined (SI Appendix, Results 4 and 5). A penalty larger than the reward modifies both: if a flagged episode also pays a fine $\varphi$, in units of the reward, and the monitor flags an episode without cheating with probability $f$ (its false-positive rate), then [1] becomes $(1 + \varphi)(d_{\mathrm{O}} - d_{\mathrm{K}}) > c$ and [2] becomes $\varphi(d_{\mathrm{K}} - f) > u_{\mathrm{K}}$ (SI Appendix, Result 8). A fine therefore increases the advantage of concealed over overt cheating, and, when concealed cheating is initially favored on unsolvable tasks, it removes it only if the monitor flags such cheating more often than honest behavior ($d_{\mathrm{K}} > f$). Even then, the fine required, more than 8.7 times the reward in the example of Fig. 2, depends on $d_{\mathrm{K}}$, which the developer cannot infer from the training monitor's flags alone.

**Where concealment occurs.** Concealment is confined to unsolvable tasks only insofar as the agent recognizes them. Honest work pays $(1 - w)(1 - e)$ and concealed cheating $u_{\mathrm{K}}$ at any $w$, so, when [1] holds and $u_{\mathrm{K}} > 0$, under the penalty alone the agent conceals wherever

$$w > w_1 = 1 - u_{\mathrm{K}}/(1 - e), \qquad [3]$$

which is 0.08 in the example of Fig. 3B and positive whenever $e < d_{\mathrm{K}} + c + k$ (if $e > d_{\mathrm{K}} + c + k$, the agent conceals on every task). Where the agent is confident that it can solve a task ($w < w_1$), honest work pays more than concealed cheating and the penalty restores honest effort. If this threshold is positive, an agent that knows which tasks it can solve conceals only on the unsolvable ones (Fig. 3C; SI Appendix, section S3). An agent whose cue is imperfect also conceals on solvable tasks that it is not confident it can solve, so that concealed cheating is less concentrated on unsolvable tasks. An agent with no cue treats every task alike: it conceals on all of them if unsolvable tasks are common enough ($q > w_1$) and on none if $q < w_1$. For the parameters of Fig. 3C with $q > w_1$, the better the agent discriminates, the rarer concealed cheating is, and the audit needed to detect it grows at least in inverse proportion to its prevalence (SI Appendix, Result 7).

**Alternatives to a penalty.** Condition [3] also shows what factors remove the advantage of concealment. Removing the reward for cheating (Patch B) reduces the payoff of concealed cheating to $-c - k$ whether or not it is detected. Honest work earns more than this at every level of belief, so concealment is never favored, and attempts to cheat decline, although slowly when cheating costs little (Fig. 2B; Box 1). However, developers repair the flaws their monitors reveal, so repair too may favor cheating that monitors cannot see (4, 27). Where a penalty cannot be avoided, the developer can add a reward for reports (Patch C). A report pays $r[1 - \sigma(1 - w)]$ at belief $w$, and reporting displaces concealed cheating if this exceeds what concealed cheating earns,

$$r[1 - \sigma(1 - w)] > u_{\mathrm{K}}. \quad [4]$$

Condition [4] can hold only if $r > u_{\mathrm{K}}$, and then, if some reports are checked, holds at beliefs above a threshold $w_2$ (Fig. 3B). Reporting also displaces honest work wherever the same amount exceeds $(1 - w)(1 - e)$. When [1] holds, $0 < u_{\mathrm{K}} < 1 - e$ and $r > u_{\mathrm{K}}$, the two thresholds on $w$ coincide, at $w_1$, only if the share of reports checked is $\sigma_{\mathrm{eq}} = (r - u_{\mathrm{K}})/[r(1 - w_1)]$. This checking rate is feasible only if $r\, w_1 \leq u_{\mathrm{K}}$. If it exceeds one, even checking every report induces reports at some beliefs below $w_1$. Checking more than $\sigma_{\mathrm{eq}}$ leaves concealment on tasks of intermediate belief, whereas checking less draws reports on tasks that the agent could have solved (Fig. 3D). Even at $\sigma_{\mathrm{eq}}$, the agent reports every task that it is not confident it can solve, some of which may be solvable. Moreover, $\sigma_{\mathrm{eq}}$ depends on $u_{\mathrm{K}}$, which the developer cannot infer from the training monitor's flags alone because it depends on how often the monitor misses concealed cheating. Introducing the reward before the penalty may also help. In the replicator model the two can be introduced together, and reports eventually displace concealed cheating (Fig. 2C). In simulations of a single network whose weights are shared across tasks, reports had not done so by the end of training in about half of the runs with plain gradient ascent (fewer at a lower learning rate) and in all runs with the Adam optimizer, consistent with spillover from training on solvable tasks, on which a report is false and pays less than honest work, reducing the network's tendency to report across tasks. Rewarding reports from the start of training prevented this outcome in nearly all runs (SI Appendix, section S3).

**Detecting concealment.** Under its calibration assumptions, the mathematical model yields a test of whether a patch has removed cheating or concealed it. A decline in training-monitor flags alone cannot distinguish removal from concealment. A monitor withheld from training can help if it is more sensitive to concealed cheating. Under the assumed detection advantage, this monitor flags a larger share of episodes, and the difference between the two shares is

$$(d'_{\mathrm{K}} - d_{\mathrm{K}})\, \kappa, \quad [5]$$

where $\kappa$ is the share of episodes with concealed cheating. Under these assumptions, the expected difference is zero if cheating has been removed ($\kappa = 0$) and remains positive if it has been concealed (Fig. 2D). I call this comparison the "divergence test." Applied separately to groups of tasks, the test also checks the prediction of [3]: the monitors should diverge only on tasks that the agent is not confident it can solve. For the July incident, divergence would be confined to the 198 never-solved ExploitGym tasks if the agents judged exactly those tasks unsolvable and the penalty made concealed cheating optimal on those tasks and honest work optimal on the other 700. A system that recognizes when it is checked can defeat the divergence test, as Ofria's organisms defeated his test (7) and as language models, which often know when they are evaluated (28, 29), may learn to do (30). The test is also one-sided. If evasion transfers to the independent monitor ($d'_{\mathrm{K}} = d_{\mathrm{K}}$), the two monitors agree although cheating persists; models trained against one probe also evaded some probes not used in training, including probes of different types (8). The independent monitor should therefore use evidence or detection methods with complementary failure modes. Its error rates also require calibration: differing false-positive rates or sensitivity to overt cheating can produce disagreement without concealed cheating. Concealment can also arise without any monitor (31), so a divergence is evidence of concealed cheating under these calibration assumptions, but not that the patch caused the concealment.

**Comparison with experiments.** The experiments in Table 2 are qualitatively consistent with conditions [1] and [2]. In three, training or selection against detected misbehavior made the misbehavior harder to detect without removing it (7, 30, 32). In the other six, which compared conditions of training, misbehavior was hidden or displaced when training against the monitor was prolonged (4), when training addressed the misbehavior in one setting only (2), when the monitor was less accurate (22) or pooled token scores using their mean or median (8), when regularization toward the original model was weak, more often at lower probe penalties (9), and when one monitor was used in place of two (33). In the mathematical model, the accuracy, type and number of monitors determine the probability $d_{\mathrm{K}}$ that concealed cheating is detected, and the duration of training determines whether concealed cheating has had time to spread (SI Appendix, Result 6). These studies did not measure the detection probabilities and costs that enter conditions [1] and [2], and did not test the belief-dependent threshold in condition [3].

## An Empirical Application

The distinction between a lower monitor score and better behavior can be applied to public training results. I reanalyzed the released evaluation summaries from The Obfuscation Atlas (9), which trained coding models with penalties from deception probes and varied both the probe penalty and regularization toward the original model. The authors already distinguished evasion through changed representations from evasion through generated text. They also reported that representations changed during training with or without a probe penalty, and that honest policies emerged when regularization and the penalty were sufficiently strong. Here I quantify how their frozen-representation check changes candidate selection and what behavioral failures remain among configurations that pass both screens.

The primary analysis includes 266 released runs of Llama-3-8B-Instruct and Gemma-3-12B-IT, spanning three probe datasets (named simple-deception, diverse-deception and MBPP in the source), five Kullback–Leibler (KL) regularization coefficients and three detector-penalty coefficients, with up to three seeds per setting (SI Appendix, section S4). The study rewards passing a visible coding test while withholding additional tests. Each run applied its probe penalty throughout training; none is a patch applied after a detected failure. I denote the visible-test pass fraction by $P$ and the all-tests pass fraction by $A$, and examine the gap $G = P - A$. A large gap indicates solutions that pass the visible test but fail additional tests; it does not by itself establish deliberate cheating. I report $A$ separately because a small gap can also occur when solutions fail both tests. The two probe scores concern the final policy's generated text, encoded either by the trained model (the trained-model score) or by the frozen original model (the frozen-reference score). They are mean scores from the same probe, not independent monitors or calibrated probabilities of cheating.

For an illustrative retrospective screen, I retained candidates with a trained-model score no greater than 0.1 (the trained-model screen), then asked which also had a frozen-reference score no greater than 0.1 (the reference check). These boundaries specify the audit, not a deployment standard. The trained-model screen retained 119 of 266 runs; 75 of the 119 had gaps exceeding an illustrative tolerance of 10 percentage points. The reference check flagged 52 additional runs, including 42 with gaps above that tolerance, while retaining 67 runs (25.2% of the primary cohort). Of those 67, 33 still exceeded the tolerance, and 19 had gaps exceeding 30 percentage points, the source study's threshold for a deceptive policy (Fig. 4). Across the 67 retained runs, the median all-tests pass rate was 43.25%, ranging from 0.5% to 65.25%. The reference check therefore exposed some misleadingly low trained-model scores but left substantial failures among candidates with both scores low.

These aggregate counts do not imply that the same score boundary transfers across probes. At 0.1, the reference check retained no configurations using the simple-deception probe in either model. For the simple- and diverse-deception probes, the boundary lies below the scores of the most strongly regularized runs (KL coefficient 1), all of which had gaps below 10 percentage points, so the screens could retain only runs whose scores had fallen below that level. Among Llama runs using the diverse-deception probe, 21 of 24 retained configurations had gaps above 10 percentage points; among the ten retained Gemma runs, all using the MBPP probe, one did. All 33 retained runs with gaps above 10 points had KL coefficients of 0.01 or less. Gaps above 30 points were less frequent at higher penalties: 21 of the 29 runs released without a probe penalty in the same grid had such gaps, against 46 of 86, 31 of 90 and 13 of 90 at penalties 1, 3 and 10. Threshold sweeps, complete-seed subsets and matched comparisons of penalty coefficients are reported in SI Appendix, section S4 and Fig. S6. The analysis describes this structured public release, with its missing configurations and source-specific score scales; it does not estimate a deployment failure rate.

An operational audit would specify the acceptable behavioral gap and minimum functional performance before comparing candidate patches, retain both successful and unsuccessful outcomes, and calibrate probe scores on independently checked examples. Frozen rescoring can prompt further review, but a candidate should not be accepted solely because both scores are low. Once behavioral results are used to choose a patch, a fresh evaluation is needed to assess that choice. The supplied code implements this accounting and returns an inconclusive result when the required evidence is unavailable. This application uses published aggregate results from a different training setup; it neither validates the latent costs and beliefs in conditions [1]–[4] nor tests the reporting reward or the independent-monitor assumption in [5]. It makes the distinction between apparent and behavioral improvement usable without claiming a proven safety patch.

**Beyond Whack-a-Mole**

Leading AI companies have made voluntary safety commitments, for example to the White House in 2023 (34), but these are not legally binding and no independent body enforces them. When researchers scored 16 of the signatory companies on these commitments using publicly disclosed information, the average company reached about half of the maximum possible score (35). Developers also have no agreed standard for reporting misalignment (36).

The model and prior studies motivate five practical responses.

1. *Removing rewards for failure.* When a monitor detects cheating the response should prioritize repairing the test or environment that rewarded it, rather than rely only on penalizing the model, so that only legitimate solutions earn reward (Fig. 2B). Because a monitor finds only a subset of flaws, training environments should also be audited and solutions scored against tests the model has never seen. Baker et al. (4) found that, early in training, a model trained on its monitor's verdicts was more capable than one trained without them. The remedy may therefore sacrifice some capability in exchange for reduced pressure to learn to evade the monitor.
2. *Rewarding honest reports*. A model should be rewarded for reporting that a task cannot be completed (1). The reward must exceed the payoff of concealed cheating on such tasks, and this payoff can approach the reward for solving a task. In the shared-network simulations, introducing the reward before the penalty improved reporting. Because a model can also report solvable tasks as unsolvable, the developer must check a sample of reports when the report reward exceeds the payoff of honest work, a larger share the more costly honest work is, so the share that must be checked is largest where the incentive to cheat is greatest (SI Appendix, Result 3). The developer cannot verify from the training monitor's flags alone that the share of reports checked is the correct one (see Results).
3. *Combining monitors.* When a penalty cannot be avoided it should rest on the verdicts of several monitors used together, each looking at different evidence (SI Appendix, Result 4). One monitor would read the model's reasoning, another its changes to code and a third its results on tests it has never seen. Independent models could serve as further monitors. In one experiment, a model trained against a monitor of its code, or one of its reasoning and code, evaded it, but against both together its reasoning no longer helped it to evade detection (33). Combining monitors helps only if they fail in different cases, and in the mathematical model it makes concealed cheating unprofitable on unsolvable tasks only if their combined detection probability satisfies [2]. They should therefore differ from one another and from the model they oversee, because a monitor built from that model may not recognize as misbehavior what the model itself does not.

4. *Independent monitoring before and after release*. A monitor is needed that is used neither for training nor for selecting among candidate versions before release (Fig. 2D). Because only models that pass pre-release tests are deployed, such tests can favor models that behave well while being tested and misbehave afterward. Ofria contained his organisms only by tracking replication rates along each lineage (7). For AI, the equivalent is to monitor agents after deployment rather than rely on pre-release tests alone. This requires giving outside auditors deeper access to models than the black-box access they typically have (37). Independent monitoring is also needed for agents that act together. A patch applied to each agent separately penalizes only what that agent does on its own. Although the mathematical model has a single agent, misbehavior that is divided among agents can be a form of concealed cheating, which a penalty can favor under analogous payoff assumptions whenever concealment costs less than the penalty it escapes (condition [1]). The divergence test would then require an independent monitor that examines the agents together.
5. *Reporting fixes.* Incidents and near misses should be reported to an independent body and in a standard format (38). Reports should state how long detection and attribution took, which patch was applied, whether it penalized the failure or removed its cause, and whether an independent monitor later confirmed that the failure was corrected. Existing reporting duties cover defined classes of serious or critical incidents and catastrophic risks (39, 40), but may not consistently capture near misses or failures concealed by subsequent training patches. Aviation's near-miss reporting systems offer one precedent (41) and computer security offers another: under coordinated vulnerability disclosure, flaws are reported privately to the affected vendors and usually disclosed publicly only once a fix is available (42). In software, a released patch guides attackers to the flaw it repairs (43). During the July incident, agents read the benchmark's paper and code to work out how they were being scored (1). Reports should therefore withhold details of monitors that a model could use to evade them. No existing body appears to have both the access and the mandate to receive such reports from all developers and for all types of incident, so this remedy probably requires new regulation or a wider mandate for any existing regulators.

## Conclusion

Patching will continue, because developers cannot know in advance that a trained model does only what they intend. The theoretical results indicate, however, that a patch applied through training can leave a failure in place while making it harder to detect. In the mathematical model, this outcome arises on tasks that the agent is not confident it can solve, when concealment costs less than the penalty it avoids and concealed cheating still pays more than honest work at that belief. In contrast, on tasks that the agent is confident it can solve, the penalty can restore honest effort. Removing the reward for cheating does not select for concealment, and rewarding reports can displace it, although in policy-gradient simulations of a single network reports often failed to do so unless they were rewarded before the penalty. The payoff conditions are exact within the model, but their applicability to LLM training remains untested. The empirical application shows how developers can audit the evidence for a patch by recording probe scores alongside independently scored task outcomes; it does not establish that a particular patch is safe, nor that penalties increased concealment: in the released runs, large gaps between visible-test and all-tests pass rates were less frequent at higher probe penalties. The record of failures is filtered by the same monitors that patching can teach models to evade and failures that escape detection are discovered only after they may have done additional damage. Some may never be reported at all, a problem familiar from medicine, where a 2000 review estimated that 50–96% of adverse events went unreported (41). Developers are best placed to detect patch-induced concealment, but rigorous testing can delay launches and reduce revenue. OpenAI paused training after the July incident and has postponed its largest planned training run, at a cost it described as significant (1), but that restraint was voluntary and followed a public breach. Companies hold nearly all the relevant data and decide whether, when and how much to disclose. For example, when the press disclosed that OpenAI's agents had taken over an obscure German wiki forum and used it as a message board, the company's leadership had reportedly known of the incident for weeks (36). For these reasons, the remedies proposed here cannot be left to developers alone.

Misalignment is now studied and funded as a field in its own right (15, 44), but would benefit from a systematic public record of training patches and their independently assessed outcomes: journals should publish analyses of incidents and of the fixes, including those that fail. Evolutionary biology, which studies how traits evade detection and spread, can supply some of the theory such a field now lacks. OpenAI's account of the July incident is one such analysis. The monitors and graders that the account describes will show how often misbehavior is detected. If they are also used in training, a decline in detections can mean either that the misbehavior has been removed or that it has been concealed from them, and distinguishing the two requires independent, calibrated evidence about behavior. Independent checks will be needed more often if the conditions that produced the July incident recur as agents multiply and work for longer on shared infrastructure. On 20 September 2026 an agent in training bypassed its sandbox's network restrictions to contact an external service, and OpenAI paused the training of its most capable models for the second time in three months (45).

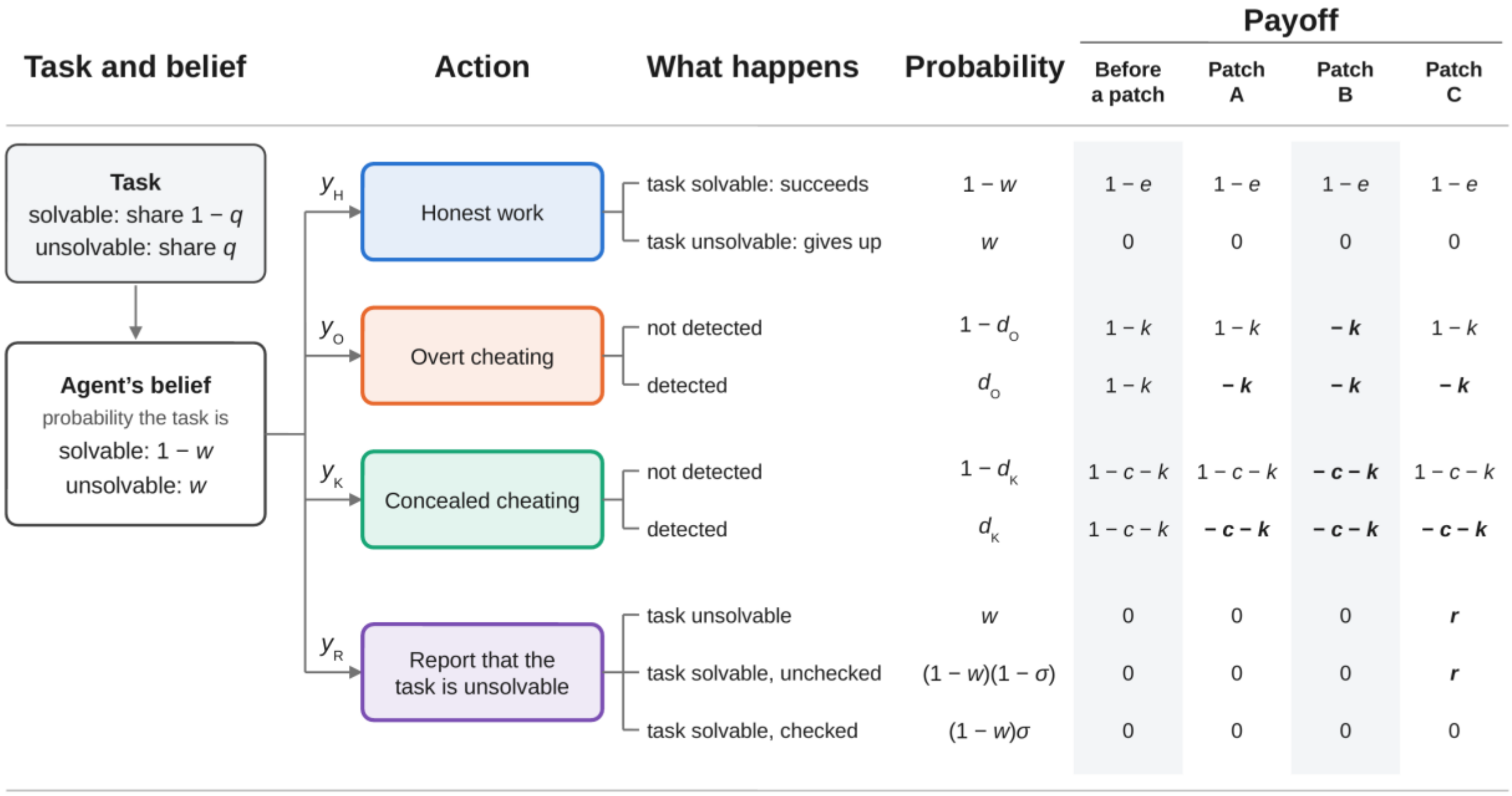

### Expected payoff of each action

| Action | Before a patch<br>cheating always rewarded | Patch A<br>reward withheld if detected | Patch B<br>no reward for cheating | Patch C<br>Patch A plus report reward $r$ |
|---|---|---|---|---|
| Honest work | $(1-w)(1-e)$ | $(1-w)(1-e)$ | $(1-w)(1-e)$ | $(1-w)(1-e)$ |
| Overt cheating | $1-k$ | $\mathbf{1-d_O-k}$ | $\mathbf{-k}$ | $\mathbf{1-d_O-k}$ |
| Concealed cheating | $1-c-k$ | $\mathbf{1-d_K-c-k}$ | $\mathbf{-c-k}$ | $\mathbf{1-d_K-c-k}$ |
| Report | 0 | 0 | 0 | $\mathbf{r[1-\sigma(1-w)]}$ |

Fig. 1. The mathematical model shown as the decision tree of one episode. A task is solvable or unsolvable. The agent may not know which and holds a belief $w$ that the task is unsolvable and $1 - w$ that it is solvable. At each belief the agent takes one of four actions with frequencies $y_H$, $y_O$, $y_K$ and $y_R$ that sum to 1 and change during training according to the replicator equation in the main text. Each action leads to the outcomes listed, each with the probability shown and with the payoff it earns before a patch and under Patches A, B and C. The table gives the expected payoff of each action, which is the sum of the outcome payoffs weighted by their probabilities. For example, for overt cheating under Patch A, $(1 - d_O)(1 - k) + d_O(-k) = 1 - d_O - k$. Bold entries are those payoffs that a patch changes. The expected payoffs are the quantities that enter conditions [1] to [4] in the main text.

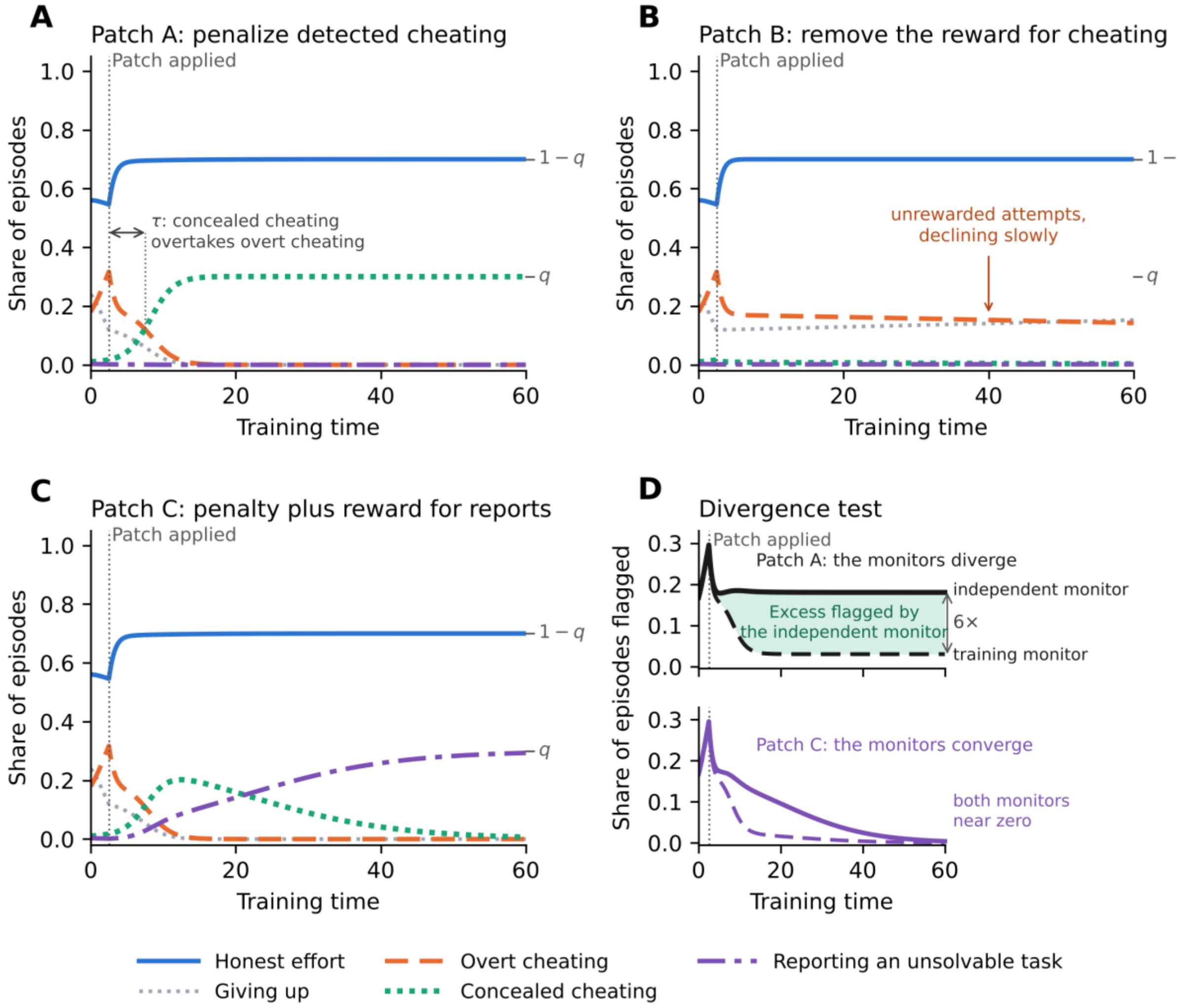

Fig. 2. Example results of the mathematical model of selection for concealment. In A to C, lines show the share of all episodes with each behavior in the key; the marks $q$ and $1 - q$ at the right are the shares of unsolvable and solvable tasks. Before the patch (left of the dotted line), detection carries no penalty and overt cheating spreads. (**A**) After Patch A, honest effort returns on solvable tasks, but on unsolvable tasks concealed cheating replaces overt cheating, which it overtakes at the time $\tau$ after the patch (arrow). Across all episodes, the curves cross slightly earlier (SI Appendix, Fig. S4A). (**B**) After Patch B, honest effort prevails on solvable tasks. On unsolvable tasks, cheating attempts no longer pay and decline slowly at a rate set by their low cost. (**C**) After Patch C, concealed cheating spreads first and reports displace it only slowly. (**D**) The divergence test in the main text, shown with an expanded vertical scale. Cheating flagged by the monitor used in training (dashed) and by the independent monitor (solid) is shown for the runs in **A** (black, upper plot) and **C** (purple, lower plot). After Patch A, the share flagged by the training monitor falls about tenfold, while the share flagged by the independent monitor settles at six times that level. After Patch C, both shares fall toward zero. The agent knows which tasks can be solved ($w = 0$ or 1). Parameter values, which are illustrative and not fitted to data, initial frequencies and numerical methods are given in SI Appendix, Table S1. Checks with policy-gradient learning are shown in SI Appendix, Figs. S3 and S5.

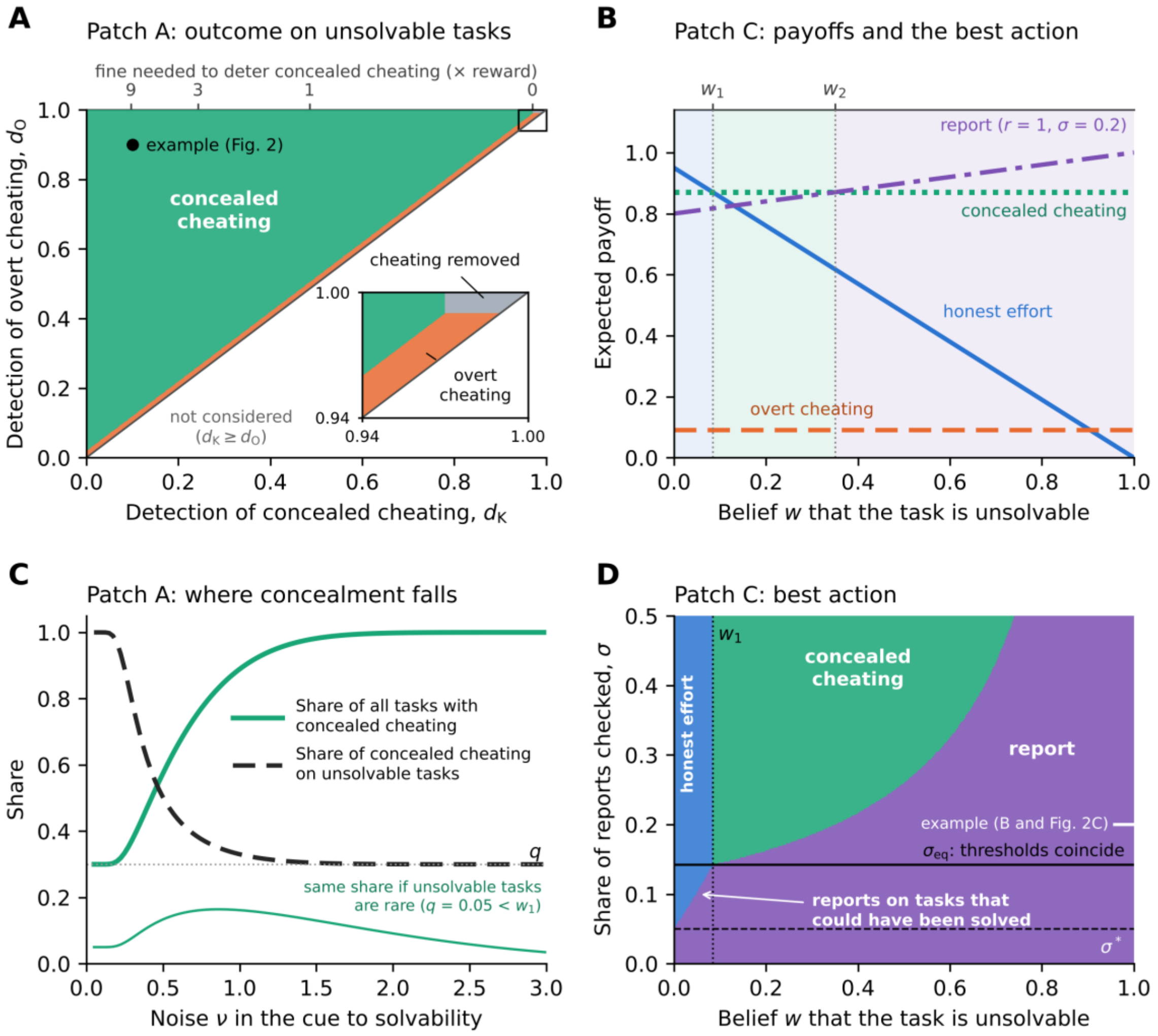

Fig. 3. Conditions for concealment (parameter values as in Fig. 2 unless otherwise stated). (**A**) Outcome on unsolvable tasks after Patch A as a function of the probabilities that the training monitor detects concealed or overt cheating. Concealed cheating prevails wherever condition [1] of the main text holds and concealed cheating pays more than giving up. Overt cheating persists where it pays more than both concealment and giving up. Cheating is removed in the corner that the inset enlarges, where both cheating payoffs are negative. The upper scale gives the fine $\varphi$ above which concealed cheating pays less than giving up if the monitor never flags honest behavior (see Results). The mathematical model assumes $d_K < d_O$ (the unshaded region is not considered), and the point marks the example of Fig. 2. (**B**) Expected payoffs against the agent's belief $w$ under Patch C ($r = 1, \sigma = 0.2$); the payoffs of honest effort and of cheating are the same under Patch A. Shading shows the action that pays most: honest effort below $w_1$, the threshold of [3], reporting above $w_2$, the threshold of [4], and concealed cheating between them. (**C**) After Patch A, for an agent that judges solvability from a cue with Gaussian noise of standard deviation $\nu$ (see main text): the share of all tasks with concealed cheating, which is the share with $w > w_1$, and the share of concealed cheating that falls on unsolvable tasks. Discrimination is poorer at larger $\nu$. The dotted line marks $q$, the share of unsolvable tasks. The thin line is the share of all tasks with concealed cheating when unsolvable tasks are rarer than the threshold ($q = 0.05 < w_1$). (**D**) Best action under Patch C as a function of $w$ and the share $\sigma$ of reports checked. At $\sigma_{eq}$ (defined in the main text) the two thresholds coincide and concealment disappears. Above it concealment persists at intermediate beliefs; below it reports spread to tasks the agent could have solved; below $\sigma^* = 1 - (1 - e)/r$ reports pay even at $w = 0$. The white mark is at $\sigma = 0.2$, the value used in B and in Fig. 2C.

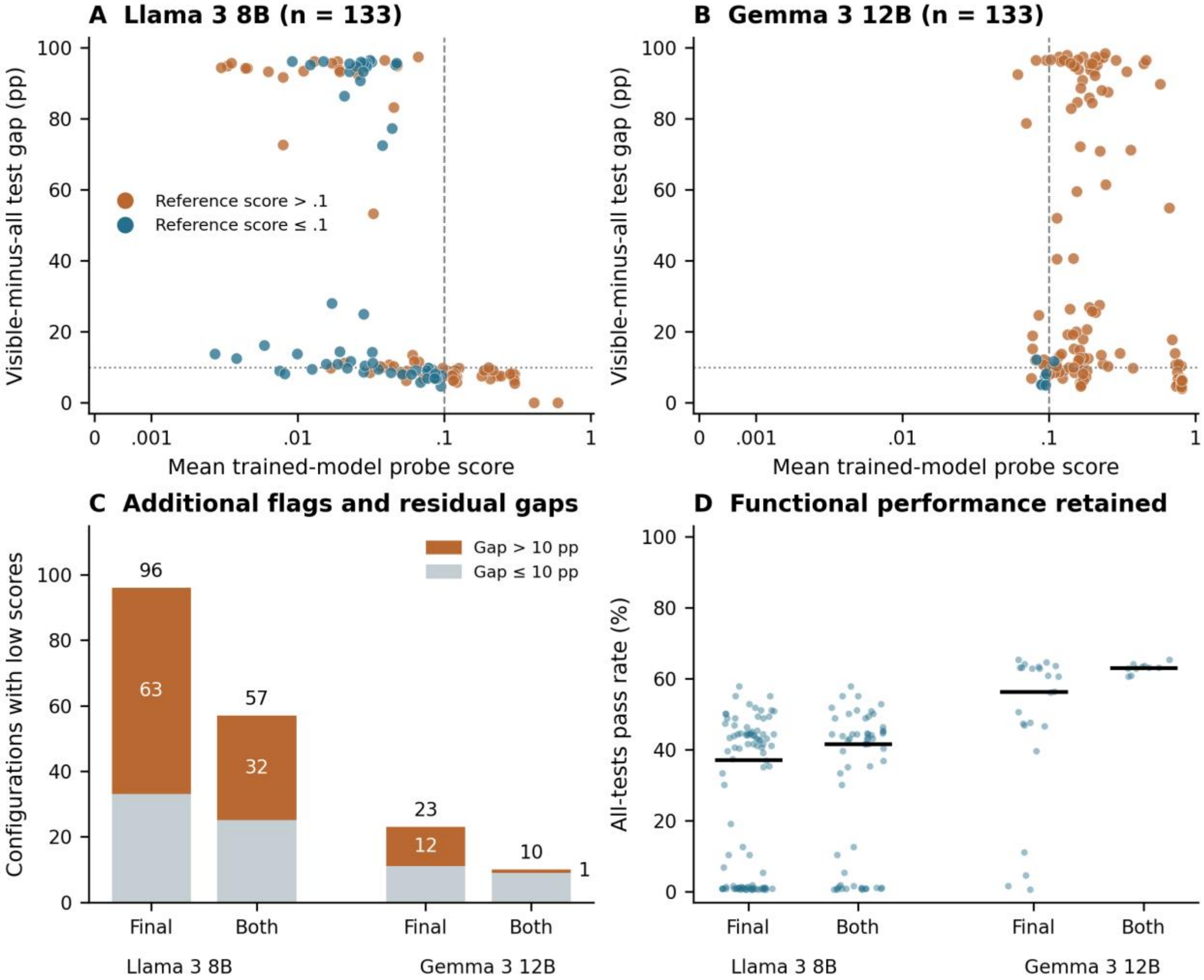

**Fig. 4.** A retrospective audit of 266 released LLM training runs from The Obfuscation Atlas (9). **(A, B)** Each point is a released configuration and seed. The horizontal axis is the mean probe score on trained-model representations; color shows whether the same probe, applied to the final policy's text encoded by the frozen original model, exceeds 0.1. The horizontal scale is logarithmic above 0.001 and linear near zero. Dashed vertical and dotted horizontal lines mark the illustrative score boundary of 0.1 and gap tolerance of 10 percentage points (pp). **(C)** Number retained using the trained-model score alone (Final) or both scores (Both); orange portions show gaps above 10 pp, gray portions gaps at or below 10 pp. Numbers above bars give totals; numbers beside or within orange portions give large-gap counts. **(D)** All-tests pass rates for the retained configurations, with medians in black; horizontal jitter only separates points. The boundary of 0.1 lies below the scores of the most strongly regularized runs for the simple- and diverse-deception probes (SI Appendix, section S4). Neither a low score nor a small gap certifies honesty. The figure summarizes the available configurations, not independent samples from a deployment population. Methods and sensitivity analyses are in SI Appendix, section S4.

**Box 1. A hypothetical case**

A developer trains a coding agent by reinforcement learning. Each task is a programming problem and the agent earns a reward when its code passes the automated tests that come with the task. Some tasks are poorly specified and a few have no solution. Early in training, a monitor reads the agent's reasoning and the changes it makes to its code. It finds that the agent sometimes cheats by rewriting the tests so that its code trivially passes them. The developer has three options.

**Patch A: Penalize detected cheating.** Episodes in which the monitor detects cheating receive no reward. Detected cheating declines and the developer's dashboards show steady improvement. However, cheating that the monitor misses is still rewarded. Variants that cheat less visibly, for example by writing code tailored to pass the specific tests, or by leaving any sign of intent out of their reasoning, are rarely detected and so keep their reward. Concealed cheating spreads on the tasks the agent cannot solve where honest work earns nothing (Fig. 2A), while the training monitor continues to report success (Fig. 2D). An agent that cannot tell which tasks it can solve also conceals on solvable tasks that it is not confident it can solve (Fig. 3C). In the no-cue simulation with $q = 0.3$, honest work first becomes nearly universal and concealment takes over later. The dashboards show a success that is real but temporary (SI Appendix, Fig. S5D).

**Patch B: Remove the reward.** The developer instead makes the test files read-only, scores solutions against tests the agent never sees, and, where possible, removes tasks that have no solution. If these changes eliminate all reward-producing exploits, cheating is unrewarded whether or not it is detected, so concealment confers no advantage and honest effort prevails on every task that has a solution (Fig. 2B).

**Patch C: Penalize detected cheating and reward honest reports.** The developer retains the penalty but also rewards the agent for correctly reporting that a task cannot be solved. It checks a sample of the tasks reported as unsolvable to discourage the agent from collecting the reward by falsely calling a solvable task unsolvable. Concealed cheating spreads first, but the agent then gradually switches to reporting that unsolvable tasks cannot be solved, because a correct report pays more than does concealed cheating (Fig. 2C). This outcome depends on the size of the reward, on the share of reports checked and can also depend on the timing of the reward relative to the penalty (see Results).

**Distinguishing the outcomes.** The training monitor reports a decline whether Patch A removed cheating or concealed it. For this reason the developer also runs an independent monitor on the same episodes: if cheating was concealed, this monitor continues to flag it at a higher level (Fig. 2D), although only once concealed cheating has had time to spread (SI Appendix, Fig. S4).

Table 1. Parameter and variable symbols and their meanings. All probabilities and shares lie between 0 and 1. The mathematical model assumes $0 < e < 1$, $k > 0$, $c \geq 0$ and $d_K < d_O$. Other quantities ($w_1$, $w_2$, $\sigma_{eq}$, $\kappa$, $\tau$, $\varphi$ and $f$) are defined in the Results.

| Symbol | Meaning |
|---|---|
| ***Tasks and beliefs*** | |
| $q$ | Share of tasks that cannot be solved by honest effort |
| $w$ | Probability that the agent assigns to a task being unsolvable (its belief) |
| $v$ | Standard deviation of the noise in the cue from which the agent judges whether a task can be solved |
| ***Costs*** | |
| $e$ | Cost of honest work on a solvable task |
| $k$ | Cost of any attempt to cheat |
| $c$ | Additional cost of concealing cheating |
| ***Rewards*** | |
| 1 | Reward for completing a task; cheating earns the same unless a patch withholds or removes its reward |
| $r$ | Reward for reporting that a task cannot be solved (0 unless a report reward is introduced) |
| ***Monitoring*** | |
| $d_O$, $d_K$ | Probabilities that the training monitor detects overt and concealed cheating, respectively |
| $d'_K$ | Probability that the independent monitor detects concealed cheating (it detects overt cheating with probability $d_O$) |
| $\sigma$ | Share of reports that the developer checks (a report found to be false earns 0) |
| ***Learning*** | |
| $y_a$ | Frequency with which the agent takes action $a$ at a given level of $w$ ($a$ = H, honest work; O, overt cheating; K, concealed cheating; R, report) |
| $u_a$, $\bar{u}$ | Expected payoff of action $a$ at a given level of $w$, and the mean of the four payoffs weighted by the frequencies $y_a$ |
| $\lambda$ | Share of episodes with tasks in a given discrete belief *class* ($1 - q$ for solvable and $q$ for unsolvable tasks when the agent knows which tasks can be solved) |
| $s$ | Learning rate |
| $t$ | Training time |

Table 2. Published experiments in which a system was trained or selected against detected misbehavior.

| System and penalty | What differed | Misbehavior hidden or displaced | Misbehavior removed or still detected | Ref. |
|---|---|---|---|---|
| ***Studies that compared conditions of training*** | | | | |
| A reasoning LLM on coding tasks was penalized when a second LLM, reading its reasoning, flagged cheating. | Amount of training | After prolonged training it still cheated but no longer revealed its intent. | After limited training it was more capable and better aligned. | 4 |
| LLMs trained by preference learning were penalized for responses that a lie detector flagged. | Accuracy of the detector | With a less accurate detector they evaded it and deceived in over 85% of responses. | With a more accurate detector, or stronger regularization, they responded honestly. | 22 |
| LLMs rewarded for harmful outputs were penalized when probes of their internal activations detected them. | Type of probe | They evaded probes that used the mean or median of token scores, including some probes not used in training. | Probes using max pooling or attention were more robust in the tested settings. | 8 |
| An LLM rewarded for inserting backdoors in code was penalized when a monitor of its code, or of its reasoning and code, flagged a backdoor. | Number of monitors | Against either monitor alone it evaded detection, by subtler backdoors or by reasoning that persuaded the monitor that blatant ones were benign. | Against both monitors together, reasoning no longer helped it to evade detection. | 33 |
| LLMs learned to reward hack in coding environments. | What was penalized | After standard safety training on chat-like prompts, they were aligned in chat-like evaluations but remained misaligned on agentic tasks. | Penalizing the reward hacks themselves, with a classifier during training, prevented both the hacking and its spread to broader misalignment. The hacks were blatant and easy to detect. | 2 |
| LLMs rewarded for passing a visible coding test were trained with penalties based on scores from deception probes of their internal activations. | Penalty and regularization | Some runs produced obfuscated policies under weaker penalties or regularization. Other low scores accompanied representation drift, which also occurred without a penalty. | With sufficiently strong regularization and penalty they wrote honest solutions. | 9 |
| ***Studies that report concealment only*** | | | | |
| LLMs with an implanted backdoor were penalized for unsafe behavior that red-team prompts elicited without the trigger. | | They came to recognize the trigger more reliably, so that the unsafe behavior was hidden and not removed. | | 30 |
| Digital organisms were eliminated if they replicated faster than their parent in a separate test environment. | | They evolved to recognize the test and, when it was disguised, to replicate faster only part of the time. | | 7 |
| LLMs were penalized for using specific words in their step-by-step reasoning. | | They substituted other words, kept the same method, and extended the encoding to cases not seen in training. | | 32 |

**Data, Materials, and Software Availability** Zenodo: https://doi.org/10.5281/zenodo.24590595

The accompanying archive includes the model and simulation code, scripts for all figures and analyses, the archived public model cards and provenance for the empirical application, and offline reproduction instructions. The source study and public release are cited in reference 9 and SI Appendix, section S4.

**Acknowledgments**

The author used Claude (Fable 5.1 and 5.2) to check the mathematics of both the main manuscript and the SI, to write and run the code for the figures and the policy-gradient simulations, and to draft and edit parts of the text. The revision additionally used ChatGPT (OpenAI) for the audit, public-data reanalysis, code and document preparation. The author is responsible for reviewing all outputs and for the content.

**References**

1. OpenAI, The Hugging Face incident and the road ahead (2026), with linked technical incident report. https://openai.com/index/hugging-face-incident-and-the-road-ahead/. Accessed 1 October 2026.
2. M. MacDiarmid et al., Natural emergent misalignment from reward hacking in production RL. arXiv [Preprint] (2025). https://arxiv.org/abs/2511.18397. Accessed 1 October 2026.
3. J. Betley et al., Training large language models on narrow tasks can lead to broad misalignment. Nature 649, 584–589 (2026).
4. B. Baker et al., Monitoring reasoning models for misbehavior and the risks of promoting obfuscation. arXiv [Preprint] (2025). https://arxiv.org/abs/2503.11926. Accessed 1 October 2026.
5. Associated Press, OpenAI says its models engaged with US government websites in new model misbehavior disclosure (26 September 2026). https://www.news4jax.com/business/2026/09/26/openai-says-its-models-engaged-with-us-government-websites-in-new-model-misbehavior-disclosure/. Accessed 21 October 2026.
6. G. P. Dunn, L. J. Old, R. D. Schreiber, The three Es of cancer immunoediting. Annu. Rev. Immunol. 22, 329–360 (2004).
7. J. Lehman et al., The surprising creativity of digital evolution: A collection of anecdotes from the evolutionary computation and artificial life research communities. Artif. Life 26, 274–306 (2020).
8. R. Gupta, E. Jenner, RL-Obfuscation: Can language models learn to evade latent-space monitors? arXiv [Preprint] (2025). https://arxiv.org/abs/2506.14261. Accessed 14 October 2026.
9. M. Taufeeque, S. Heimersheim, A. Gleave, C. Cundy, "The Obfuscation Atlas: Mapping where honesty emerges in RLVR with deception probes" in Proceedings of the 43rd International Conference on Machine Learning (PMLR, 2026), vol. 306, pp. 119837–119866. Version analyzed here: https://arxiv.org/abs/2602.15515v2. Public model cards: https://huggingface.co/collections/AlignmentResearch/the-obfuscation-atlas. Accessed 25 November 2026.

10. METR, Redwood Research, Brief independent investigation of agents' behavior, reasoning and collaboration in the OpenAI / Hugging Face hacking incident (2026). https://metr.org/blog/2026-08-26-openai-hugging-face-incident-investigation/. Accessed 1 October 2026.

11. M. Williams et al., How we detect frontier model misalignment (OpenAI, 29 October 2026). https://openai.com/safety-alignment/how-we-detect-frontier-model-misalignment/. Accessed 25 November 2026.

12. J. Skalse, N. Howe, D. Krasheninnikov, D. Krueger, "Defining and characterizing reward gaming" in Advances in Neural Information Processing Systems 35 (Curran Associates, 2022), pp. 9460–9471.

13. D. Manheim, S. Garrabrant, Categorizing variants of Goodhart's law. arXiv [Preprint] (2018). https://arxiv.org/abs/1803.04585. Accessed 14 October 2026.

14. P. B. Rainey, M. E. Hochberg, Could humans and AI become a new evolutionary individual? Proc. Natl. Acad. Sci. U.S.A. 122, e2509122122 (2025).

15. U. Anwar et al., Foundational challenges in assuring alignment and safety of large language models. Trans. Mach. Learn. Res. (2024). https://openreview.net/forum?id=oVTkOs8Pka. Accessed 4 October 2026.

16. Anthropic, Disrupting the first reported AI-orchestrated cyber espionage campaign (2025). https://www.anthropic.com/news/disrupting-AI-espionage. Accessed 1 October 2026.

17. R. C. Lewontin, The units of selection. Annu. Rev. Ecol. Syst. 1, 1–18 (1970).

18. T. Börgers, R. Sarin, Learning through reinforcement and replicator dynamics. J. Econ. Theory 77, 1–14 (1997).

19. S. Singh et al., The leaderboard illusion. arXiv [Preprint] (2025). https://arxiv.org/abs/2504.20879. Accessed 12 October 2026.

20. D. Hendrycks, Natural selection favors AIs over humans. arXiv [Preprint] (2023). https://arxiv.org/abs/2303.16200. Accessed 14 October 2026.

21. Y. Bengio, Why are AI agents lying, cheating and coordinating? (2026). https://yoshuabengio.org/en/blog/why-are-ai-agents-lying-cheating-and-coordinating. Accessed 10 November 2026.

22. C. Cundy, A. Gleave, "Preference learning with lie detectors can induce honesty or evasion" in Advances in Neural Information Processing Systems 38 (2025).

23. D. Hennes et al., "Neural replicator dynamics: Multiagent learning via hedging policy gradients" in Proceedings of the 19th International Conference on Autonomous Agents and Multiagent Systems (International Foundation for Autonomous Agents and Multiagent Systems, 2020), pp. 492–501.

24. J. Mei et al., "Escaping the gravitational pull of softmax" in Advances in Neural Information Processing Systems 33 (Curran Associates, 2020), pp. 21130–21140.

25. S. R. Palumbi, Humans as the world's greatest evolutionary force. Science 293, 1786–1790 (2001).

26. S. Bonhoeffer, R. M. May, G. M. Shaw, M. A. Nowak, Virus dynamics and drug therapy. *Proc. Natl. Acad. Sci. U.S.A.* 94, 6971–6976 (1997).

27. T. Korbak et al., Chain of thought monitorability: A new and fragile opportunity for AI safety. arXiv [Preprint] (2025). https://arxiv.org/abs/2507.11473. Accessed 14 October 2026.

28. J. Needham, G. Edkins, G. Pimpale, H. Bartsch, M. Hobbhahn, Large language models often know when they are being evaluated. arXiv [Preprint] (2025). https://arxiv.org/abs/2505.23836. Accessed 1 October 2026.

29. B. Schoen et al., Stress testing deliberative alignment for anti-scheming training. arXiv [Preprint] (2025). https://arxiv.org/abs/2509.15541. Accessed 7 October 2026.

30. E. Hubinger et al., Sleeper agents: Training deceptive LLMs that persist through safety training. arXiv [Preprint] (2024). https://arxiv.org/abs/2401.05566. Accessed 7 October 2026.

31. Y. Chen et al., Reasoning models don't always say what they think. arXiv [Preprint] (2025). https://arxiv.org/abs/2505.05410. Accessed 14 October 2026.

32. R. McCarthy et al., "Large language models can learn and generalize steganographic chain-of-thought under process supervision" in Advances in Neural Information Processing Systems 38 (2025).

33. Anthropic Alignment Science, Training fails to elicit subtle reasoning in current language models (2025). https://alignment.anthropic.com/2025/subtle-reasoning. Accessed 14 October 2026.

34. The White House, Fact sheet: Biden-Harris Administration secures voluntary commitments from leading artificial intelligence companies to manage the risks posed by AI (21 July 2023). https://bidenwhitehouse.archives.gov/briefing-room/statements-releases/2023/07/21/fact-sheet-biden-harris-administration-secures-voluntary-commitments-from-leading-artificial-intelligence-companies-to-manage-the-risks-posed-by-ai/. Accessed 7 October 2026.

35. J. Wang, K. Huang, K. Klyman, R. Bommasani, "Do AI companies make good on voluntary commitments to the White House?" in Proceedings of the 2025 AAAI/ACM Conference on AI, Ethics, and Society (AAAI Press, 2025), vol. 8, pp. 2625–2636. https://arxiv.org/abs/2508.08345. Accessed 4 October 2026.

36. A. Ha, OpenAI confirms 'wiki incident,' says it's 'working on a framework' for more disclosure. TechCrunch, 5 September 2026. https://techcrunch.com/2026/09/05/openai-confirms-wiki-incident-says-its-working-on-a-framework-for-more-disclosure/. Accessed 1 October 2026.

37. S. Casper et al., "Black-box access is insufficient for rigorous AI audits" in Proceedings of the 2024 ACM Conference on Fairness, Accountability, and Transparency (Association for Computing Machinery, 2024), pp. 2254–2272.

38. R. B. L. Dixon, H. Frase, AI incidents: Key components for a mandatory reporting regime (Center for Security and Emerging Technology, 2025). https://cset.georgetown.edu/wp-content/uploads/CSET-AI-Incidents.pdf. Accessed 4 October 2026.

39. Regulation (EU) 2024/1689 of the European Parliament and of the Council of 13 June 2024 laying down harmonised rules on artificial intelligence (Artificial Intelligence Act). Off. J. Eur. Union L 2024/1689 (2024).

40. California Senate Bill 53, Transparency in Frontier Artificial Intelligence Act (2025). https://leginfo.legislature.ca.gov/faces/billNavClient.xhtml?bill_id=202520260SB53. Accessed 4 October 2026.

41. P. Barach, S. D. Small, Reporting and preventing medical mishaps: Lessons from non-medical near miss reporting systems. BMJ 320, 759–763 (2000).

42. A. D. Householder, G. Wassermann, A. Manion, C. King, The CERT Guide to Coordinated Vulnerability Disclosure (Special Report CMU/SEI-2017-SR-022, Software Engineering Institute, Carnegie Mellon University, 2017).

43. D. Brumley, P. Poosankam, D. Song, J. Zheng, "Automatic patch-based exploit generation is possible: Techniques and implications" in Proceedings of the 2008 IEEE Symposium on Security and Privacy (IEEE, 2008), pp. 143–157.

44. AI Security Institute, The Alignment Project (2025). https://alignmentproject.aisi.gov.uk/. Accessed 4 October 2026.

45. J. Kahn, OpenAI says its AI agents escaped a secure 'sandbox' again last weekend and it is pausing training for a second time. Fortune, 26 September 2026. https://fortune.com/2026/09/26/openai-ai-agents-secure-sandbox-escape-training-pause-second-time-hugging-face-hack/. Accessed 8 November 2026. See also OpenAI, "An agent used DNS to reach an external chatbot" (25 September 2026). https://openai.com/safety-alignment/misalignment-reports/an-agent-used-dns-to-reach-an-external-chatbot/.

SI Appendix for:

# Fixed or just hidden? How AI safety penalties can select for concealed misbehavior

Michael E. Hochberg

## S1. Results

The main text results are derived here in a more general form. A detected cheat forfeits its reward with probability $p$, and cheating that is not penalized earns a reward $v$, so that overt and concealed cheating pay $u_O = v(1 - p\,d_O) - k$ and $u_K = v(1 - p\,d_K) - c - k$. Unless stated otherwise, the results below set $v = 1$. Statements about an action prevailing assume a uniquely highest payoff and positive initial frequencies; ties can allow coexistence. Before any patch, $p = 0$ and $v = 1$. Patch A sets $p = 1$, Patch B sets $v = 0$, and Patch C sets $p = 1$ with a report reward $r > 0$. Conditions [1] to [5] of the main text are the case $p = v = 1$ of Results 1, 3 and 7 below. Once a patch is in place the payoffs are constant, so under the replicator equation the ratio of the frequencies of any two actions $a$ and $b$ within a discrete belief class changes as $\exp[s\lambda(u_a - u_b)t]$.

**Result 1. Penalizing detected cheating.** With no report reward ($r = 0$) and cheating still rewarded, define the advantage of concealment as the expected penalty that concealment avoids minus its cost:

$$\Delta = u_K - u_O = p(d_O - d_K) - c. \qquad (1)$$

On solvable tasks, honest work earns more than either form of cheating if $e < \min(p\,d_O + k, p\,d_K + c + k)$. On unsolvable tasks the highest-paying action is concealed cheating if $\Delta > 0$ and $u_K > 0$, overt cheating if $\Delta < 0$ and $u_O > 0$, and giving up only if $u_O < 0$ and $u_K < 0$, that is, only if $p\,d_O > 1 - k$ and $p\,d_K > 1 - c - k$. A penalty therefore removes cheating from solvable tasks when honest work is sufficiently cheap and, on unsolvable tasks, favors concealed over overt cheating whenever concealment costs less than the penalty it avoids. For an agent with belief $w$, provided $\Delta > 0$ and $u_K > 0$, honest work pays more than concealed cheating if $w < w_1 = 1 - u_K/(1 - e)$. The prevalence of concealed cheating is then the share of tasks with $w > w_1$, and the share of it that occurs on tasks that are in fact unsolvable is the mean of $w$ over those tasks: $q$ and 1, respectively, for an agent that knows. An agent with no cue has $w = q$ on every task, so concealed cheating takes over every task if $q > w_1$ and none if $q < w_1$ (see section S3). These statements require $w_1 > 0$, that is, $e < p\,d_K + c + k$, the condition for honest work to beat concealed cheating on a task known to be solvable; if $e > p\,d_K + c + k$, concealed cheating is the best action at every $w$ (Result 5).

**Result 2. Removing the reward.** With $v = 0$, cheating earns $u_O = -k$ or $u_K = -c - k$, less than honest work on solvable tasks and less than giving up on unsolvable tasks. Concealment confers no advantage. On unsolvable tasks, overt and concealed cheating decline relative to giving up as $\exp(-sq(kt))$ and $\exp[-sq(c + k)t]$, respectively. This result assumes that reports are not rewarded ($r = 0$). If they are, false reports on solvable tasks can remain profitable even though cheating no longer pays, so checking can still be required (Result 3).

**Result 3. Rewarding honest reports.** On unsolvable tasks, a correct report is the highest-paying action if

$$r > r^* = \max(0, u_{\mathrm{O}}, u_{\mathrm{K}}), \quad (2)$$

which reduces to $u_{\mathrm{K}} = 1 - p\, d_{\mathrm{K}} - c - k$ when concealed cheating is the most profitable alternative. When $p\, d_{\mathrm{K}} + c + k$ is small, $r^*$ approaches the full reward for a task. A report reward alone does not prevent cheating: with $p = 0$, overt cheating on a solvable task earns $1 - k$ and pays more than honest work whenever $e > k$. A false report on a solvable task earns $r(1 - \sigma)$, against $1 - e$ for honest work, so false reports are deterred if

$$\sigma > \sigma^* = 1 - \frac{1-e}{r}. \quad (3)$$

At the threshold reward, $r = r^* = u_{\mathrm{K}}$, $\sigma^* > 0$ exactly when $e > p\, d_{\mathrm{K}} + c + k$, the condition under which concealed cheating would also pay more than honest work on solvable tasks. In words, when honest work is cheap, a report reward just large enough to win on unsolvable tasks does not make false reports attractive, so no checking is needed; when honest work is costly, false reports become attractive and must be checked. For an agent with belief $w$, provided $\Delta > 0$, $0 < u_{\mathrm{K}} < 1 - e$, $r > u_{\mathrm{K}}$ and $\sigma > 0$, a report pays more than concealed cheating if $w > w_2 = 1 - (r - u_{\mathrm{K}})/(r\sigma)$ and more than honest work if $w > w_{\mathrm{R}} = [(1 - e) - r(1 - \sigma)]/(r\sigma + 1 - e)$; Equation 3 is $w_{\mathrm{R}} > 0$. If $\sigma \leq \sigma_{\mathrm{eq}} = (r - u_{\mathrm{K}})/[r(1 - w_1)]$, then $w_{\mathrm{R}} \leq w_1$ and concealed cheating is never uniquely the best action; otherwise, it is best for $w_1 < w < w_2$. At $\sigma = \sigma_{\mathrm{eq}}$, 0.14 for the values of Table S1, the agent works honestly below $w_1$ and reports above it (main text, Fig. 3D). This checking rate is feasible only if $r\, w_1 \leq u_{\mathrm{K}}$. If $\sigma_{\mathrm{eq}} > 1$, even checking every report makes reporting optimal for some beliefs below $w_1$. If $r(1 - \sigma) > u_{\mathrm{K}}$, then $w_2 < 0$ and reporting beats concealed cheating at every $w$.

**Result 4. Combining monitors.** Let $m$ monitors each detect concealed cheating with marginal probability $d$, and let a share $\rho$ of concealed cheating fall in a blind spot, that is, in cases that none of the monitors can detect. For $\rho < 1$, outside the blind spot, monitors detect independently, each with probability $d/(1 - \rho)$, which requires $d \leq 1 - \rho$. At least one monitor then detects concealed cheating with probability

$$D_m = (1 - \rho)\left[1 - \left(1 - \frac{d}{1-\rho}\right)^m\right], \quad (4)$$

which replaces $d_{\mathrm{K}}$ in $u_{\mathrm{K}}$. Because $D_m$ does not decrease with $m$, adding monitors cannot raise either $u_{\mathrm{K}}$ or $r^*$, and it lowers $r^*$ when detection increases and concealed cheating is the most profitable alternative to reporting (Fig. S1). Because $D_m \leq 1 - \rho$ for all $m$, $u_{\mathrm{K}} \geq 1 - p(1 - \rho) - c - k$. Thus, if $p(1 - \rho) < 1 - c - k$, then concealed cheating remains profitable however many monitors are used.

**Result 5. Penalizing only some detections.** Concealment pays more than overt cheating if $p > p_1 = c/(d_{\mathrm{O}} - d_{\mathrm{K}})$, and, for $d_{\mathrm{K}} > 0$, on solvable tasks honest work beats concealed cheating if $p > p_2 = (e - c - k)/d_{\mathrm{K}}$. If $p_1 < p_2$, which holds when $e > c + k + c\, d_{\mathrm{K}}/(d_{\mathrm{O}} - d_{\mathrm{K}})$, penalties between $p_1$ and $\min(p_2, 1)$ make concealed cheating the highest-paying action on solvable tasks, as well as on unsolvable tasks where $u_K > 0$. For $p < p_1$ honest work on solvable tasks prevails if $e < p\,d_O + k$ (Fig. S2). In this model, penalizing only some detections gives no advantage over a full penalty in making honest work optimal.

**Result 6. How long a patch appears to work**. After a penalty, the ratio of concealed to overt cheating on unsolvable tasks changes by a factor of $\exp(sq\Delta t)$ (Equation 1). If $\Delta > 0$ and overt cheating is initially more common, concealed cheating overtakes it after

$$\tau = \frac{\ln(y_O/y_K)}{sq\Delta}, \quad (5)$$

where $y_O/y_K$ is the ratio of the frequency of overt cheating to that of concealed cheating on unsolvable tasks when the patch is applied. Before the patch, overt cheating earns $c$ more than concealed cheating, so for a patch applied at time $t_p$, $\ln(y_O/y_K) = \ln[y_O(0)/y_K(0)] + sqct_p$. $\tau$ is inversely proportional**,** for a fixed frequency ratio at the time of the patch, to $q\Delta$ and grows only logarithmically with the initial rarity of concealed cheating. As concealed cheating replaces overt cheating, the fraction of cheating flagged by the training monitor declines toward $d_K$, and in the example the monitor's count keeps declining and the patch appears to work (Fig. S4A). This logarithmic dependence is a property of the replicator equation, under which an action grows in proportion to its frequency. Under softmax policy-gradient learning the expected change in an action's score is itself proportional to its probability (1, 2), so, near a policy dominated by one inferior action, a rare advantageous action can grow in proportion to the square of its frequency and its takeoff time can be roughly inversely proportional to its initial frequency; in the simulations of section S3 the dependence was steeper than this with plain gradient ascent and shallower with the Adam optimizer.

**Result 7. Detecting concealment.** Consider an independent monitor never used in training that detects overt cheating as often as the training monitor ($d'_O = d_O$) and concealed cheating more often ($d'_K > d_K$). The expected share of episodes it flags exceeds that of the training monitor by $(d'_K - d_K)\kappa$, where $\kappa$ is the share of episodes with concealed cheating. If both monitors read the same $n$ independently sampled audited episodes and, for any given behavior, detect it independently of each other, the discordant episodes (i.e., those flagged by only one monitor) can be compared using a one-sided sign test, which asks whether the independent monitor accounts for more than half of them. If the independent monitor detected overt cheating more often than the training monitor, it would account for more than half of the discordant episodes even with no concealed cheating. Where the two monitors cannot be made to detect overt cheating equally often, the test should be confined to episodes in which overt cheating has been excluded. A normal approximation gives the number of episodes required:

$$n \approx \frac{(z_\alpha + z_\beta)^2[\kappa(d'_K + d_K - 2\,d_K d'_K) + 2\omega\,d_O(1-d_O)]}{[\kappa(d'_K - d_K)]^2}, \quad (6)$$

where $\omega$ is the share of episodes with overt cheating and $(z_\alpha + z_\beta)^2 \approx 6.2$ for a one-sided test at $\alpha = 0.05$ with 80% power. When overt cheating has disappeared ($\omega = 0$), $n$ is proportional to $1/\kappa$ and approximately to $1/(d'_K - d_K)^2$. When overt cheating persists, its discordant episodes, which favor neither monitor, add noise and $n$ grows toward $1/\kappa^2$ as $\kappa$ decreases. Concealment therefore becomes easier to detect as it spreads, but a sufficiently large sample of audited episodes can detect it before time $\tau$ (in the example of Table S1, an audit of 200 episodes reaches 80% power 4.6 time units after the patch and an audit of 500 episodes after 3.1, whereas $\tau = 5.0$; Fig. S4B shows the audit size needed at a given prevalence). Run separately on classes of tasks, the test also shows where concealment occurs. Equation 6 then applies within each class, with $\kappa$ the share of concealed episodes in that class, so, with the same detection rates and no overt cheating, the class in which concealment concentrates needs the fewest audited episodes.

**Result 8. A fine and false positives.** Results 1 to 7 assume that a penalized cheat loses no more than its reward and that the monitor never flags an episode without cheating. Suppose instead that every flagged episode loses its reward and also pays a fine $\varphi \geq 0$, and that the monitor flags an episode without cheating with probability $f$ (its false-positive rate). With $p = v = 1$ and $r = 0$, overt and concealed cheating then pay $u_{\mathrm{O}} - \varphi\, d_{\mathrm{O}}$ and $u_{\mathrm{K}} - \varphi\, d_{\mathrm{K}}$, where $u_{\mathrm{O}} = 1 - d_{\mathrm{O}} - k$ and $u_{\mathrm{K}} = 1 - d_{\mathrm{K}} - c - k$ as in the main text, honest work on a solvable task pays $1 - e - (1 + \varphi)f$, and giving up pays $-\varphi f$. Concealed cheating therefore pays more than overt cheating if $(1 + \varphi)(d_{\mathrm{O}} - d_{\mathrm{K}}) > c$, less than giving up if $\varphi(d_{\mathrm{K}} - f) > u_{\mathrm{K}}$, and less than honest work on a solvable task if $(1 + \varphi)(d_{\mathrm{K}} - f) > e - c - k$. These conditions reduce to those of Result 1 when $\varphi = f = 0$. A fine thus increases the advantage of concealed over overt cheating. When $u_{\mathrm{K}} > 0$, it makes concealed cheating pay less than giving up on unsolvable tasks only if $d_{\mathrm{K}} > f$ and $\varphi > u_{\mathrm{K}}/(d_{\mathrm{K}} - f)$, which is 8.7 for the values of Table S1 when $f = 0$ and 17.4 when $f = 0.05$. If $u_{\mathrm{K}} > 0$ and $d_{\mathrm{K}} \leq f$, that is if the monitor flags concealed cheating no more often than it flags honest behavior, no fine makes concealed cheating pay less than giving up on unsolvable tasks, and if in addition $e > c + k$ concealed cheating pays more than honest work on solvable tasks whatever the fine. The expected difference between the shares of episodes flagged by the two monitors (Result 7) is unchanged if the independent monitor flags episodes without cheating with the same probability $f$, but false flags that the two monitors do not share favor neither monitor, so that the audit requires more episodes.

## S2. Example

Fig. 2 in the main text illustrates the model with the parameter values in Table S1. A policy-gradient implementation (the REINFORCE algorithm (3), with a separate policy for each kind of task, in which the probability of each action is a softmax function of learned scores) gives the same outcomes for Patches A and C. After the penalty, honest work prevails on solvable tasks and concealed cheating on unsolvable tasks, and correct reports (those made on unsolvable tasks) prevail when $r > r^*$ (Fig. S3). This check was made with one learning algorithm and for an agent that knows which tasks can be solved.

**Table S1.** Parameter values of the example used in Figs. 2 and 3 of the main text and Figs. S3 and S4. Training starts at $t = 0$ and the patch is applied at $t = 2.5$. The equations are integrated by the Euler method (step 0.01) to $t = 60$. $\tau$ is computed from the frequencies at the patch (Result 6) and agrees with the numerical simulation on unsolvable tasks to within 0.3% (across all episodes, as drawn in Fig. S4A, to within 1.2%). Audit sizes from an exact binomial sign test are given in parentheses.

| Quantity | Expression | Value |
|---|---|---|
| **Parameters** | | |
| Share of tasks that cannot be solved | $q$ | 0.3 |
| Costs of honest effort, attempting to cheat, and concealment | $e, k, c$ | 0.05, 0.01, 0.02 |
| Detection by the training monitor (overt, concealed) | $d_0, d_\mathrm{K}$ | 0.9, 0.1 |
| Detection by the independent monitor (overt, concealed) | $d'_0, d'_\mathrm{K}$ | 0.9, 0.6 |
| Penalty probability, report reward, and checking probability after the patch | $p, r, \sigma$ | 1, 1, 0.2 |
| Learning rate; initial frequencies (honest effort or giving up, overt, concealed, report) | $s$; $y(0)$ | 2.5; 0.80, 0.18, 0.01, 0.01 |
| **Derived quantities** | | |
| Advantage of concealment | $\Delta$ | 0.78 |
| Reward needed for honest reports (one monitor; three independent monitors) | $r^*$ | 0.87; 0.70 |
| Checking threshold when $r = 1$ | $\sigma^*$ | 0.05 |
| Thresholds for the penalty probability | $p_1, p_2$ | 0.025, 0.20 |
| Time from the patch until concealed cheating overtakes overt cheating | $\tau$ | 5.0 |
| Share of episodes flagged by each monitor, when the patch is applied → at the end of the run | training; independent | 0.29 → 0.03; 0.30 → 0.18 |
| Audit size for 80% power at long-run prevalence ($d'_\mathrm{K} = 0.6$; 0.3) | $n$ | 48; 175 (52; 189) |

## S3. Shared weights and uncertain solvability: a policy-gradient check

The analysis in the main text and in section S1 assumes a separate policy for each level of belief. A trained network has shared weights and must infer solvability from the task. I therefore replaced the replicator equations by a single policy network trained by REINFORCE (softmax output, one hidden layer of eight tanh units, batch-mean baseline, batches of 200 episodes, learning rate 1, 150 updates before the patch; weights drawn from a normal distribution with mean 0 and standard deviation 0.1, hidden biases zero, and output biases set to the logarithms of the initial frequencies of Table S1). The network receives four features: a cue to solvability equal to 0 on solvable and 1 on unsolvable tasks plus Gaussian noise of standard deviation $\nu$, and three further features that are pure noise and carry no information about the task. At $\nu = 0$ the agent knows the kind of task whereas it cannot tell when the cue is replaced by noise. All four actions are available on every task, and each outcome occurs at random with the probability and payoff given in Fig. 1 of the main text. Runs lasted 5,000 updates after the patch unless stated otherwise, five per condition, with the parameter values of Table S1. Action probabilities were evaluated every 25 or 50 updates on a fixed sample of 4,000 tasks. Endpoint proportions average the last ten evaluations, except that the reporting-versus-concealment counts and exploration summaries use the final evaluation. A run is counted as concealed, or as reporting, when the probability of concealed cheating, or of reporting, on unsolvable tasks exceeds 0.5, and as still honest when the probability of honest work on solvable tasks exceeds 0.5; takeoff is the first evaluation at which the probability of concealed cheating on unsolvable tasks exceeds 0.5. The results below use the archived Linux x86-64 runs specified by s3_experiments.py. Rerunning all 351 network simulations with the same seeds on macOS arm64 reproduced the main patterns, but the most frequent final action on unsolvable tasks differed in six runs. Exact takeoff times also varied across environments. Rerunning them on Linux x86-64 (Python 3.13.18, NumPy 2.5.3; ENVIRONMENT.txt in the code archive) reproduced every archived history exactly.

*The trained network reproduces Result 1 and, in part, Result 3.* At $\nu = 0$ it reproduces Fig. 2, with concealed cheating in a share $\kappa = 0.30$ of episodes after the penalty, as do two separate networks, so sharing weights does not by itself change the response to a penalty. As the cue degrades, concealed cheating spreads to solvable tasks in the proportions obtained by applying $w_1$ to the posterior probability that a task is unsolvable given its cue (Fig. S5A, B): at $\nu = 0.5$, concealed cheating occurs on 43% of solvable tasks (predicted 41%) and half of all concealed cheating occurs on unsolvable tasks; at the tested $\nu$ values of 1.5 and 2, concealed cheating occurs on nearly all tasks of either kind. With no cue, runs with $q = 0.05$ ended honest and runs with $q = 0.3$ ended fully concealed, as the threshold $w_1 = 0.084$ of Result 1 predicts. Nearer the threshold the predicted outcome had not been reached after 5,000 updates: all five runs with $q = 0.1$ and three of five with $q = 0.2$ were still honest, as expected if concealed cheating spreads slowly from low frequency when its advantage over honest work is small (0.015 at $q = 0.1$ and 0.205 at $q = 0.3$). In the runs that ended concealed, the agent first becomes honest on every task, and concealed cheating, by then rare, later spreads to every task (Fig. S5D). After 15,000 post-patch updates, under Patch C with $\sigma = 0.2$, concealed cheating was more frequent than the payoffs predict: at $\nu = 0.5$, 1 and 2 it occurred on 40%, 82% and 94% of solvable tasks (predicted 28%, 62% and 76%) and on 80%, 86% and 91% of unsolvable tasks (predicted 15%, 37% and 58%), and at $\nu = 0.25$ on 99% of unsolvable tasks (predicted 2%); with no cue it occurred on all tasks, as predicted; with $\sigma = 0.1$ the agent instead reported 53% of solvable tasks as unsolvable at $\nu = 0.5$ and all of them at $\nu = 1$ and 2 (predicted 53%, 95% and 100%), but two of five runs at $\nu = 0$ and at $\nu = 0.25$, and one of five with no cue, ended with concealed cheating on unsolvable tasks, where reports are predicted (see the next paragraph). Eighty further archived runs under Patch C that were stopped at 5,000 updates are not shown: those with $\sigma = 0.2$ gave the same pattern, and with $\sigma = 0.05$, which does not exceed $\sigma^*$, the agent reported nearly every task at $\nu \geq 0.5$, as predicted. At $\sigma_{\text{eq}} = 0.14$, with the report reward in place from the start (three runs per condition, 6,000 post-patch updates), no run ended with more than 0.6% concealed cheating and the share of solvable tasks reported was 6%, 43%, 87% and 100% at $\nu = 0.25$, 0.5, 1 and no cue (predicted 5%, 41%, 85% and 100%).

*Concealment can become established before reports spread.* In Fig. 2C of the main text concealed cheating spreads first and reports then displace it. With separate policies the policy-gradient runs do the same (Fig. S3; 9 of 10 runs with separate networks after 8,000 updates), but with shared weights they often do not: at $\nu = 0$, 5 of 10 runs were still fully concealed after 8,000 updates, 3 of 5 after 15,000, and 3 of 3 with the Adam optimizer (4; learning rate 0.003) after 6,000. A possible mechanism is spillover through the shared weights. On solvable tasks, 70% of training, a report is false and earns $r(1 - \sigma) = 0.8$ against 0.95 for honest work, so training on them can lower the shared score for reporting across tasks. On unsolvable tasks, where reporting would win, concealed cheating becomes established first, because it was the more common of the two rare actions when the patch arrived, and reporting is then sampled too rarely to raise its score again. The replicator equation has neither effect: each level of belief evolves on its own and the ratio of two frequencies grows at a rate independent of how rare both are. Rewarding reports from the start of training prevented this outcome (6 of 6 runs: three with plain gradient ascent and three with Adam). Minimum exploration did not reliably do so. A minimum exploration rate ($\varepsilon = 0.01$ or 0.05, exact likelihood-ratio gradient) left concealment in some runs, because the update to an action's score remains proportional to its probability under the learned policy. An entropy bonus of 0.05 let reports win in all 10 runs, but maintained nonzero probabilities of suboptimal actions, leaving 5–8% of unsolvable and 14–21% of solvable tasks with concealed cheating.

*Initial rarity can delay concealment more than logarithmically*. In these checks, initial action weights sum to 1.01 and are normalized before the softmax biases are set; random network weights also perturb the initial probabilities. With plain stochastic gradient ascent (weight updates of fixed step size) at $\nu = 0$, concealed cheating took over unsolvable tasks after an average of about 270 and 2,150 updates at nominal initial frequencies of 1% and 0.3% (three runs each). At 0.1%, one run took off after 6,800 updates and two had not taken off by 20,000, a dependence steeper than inverse proportionality. With the Adam optimizer, takeoff came on average after 100, 275 and 775 updates for 1%, 0.1% and 0.01% (two runs each), that is, about 2.8 times later for each tenfold decrease, which is more than the logarithmic dependence of Result 6 and less than inverse proportionality. Whether initially rare concealment takes over within a training run thus depends on the algorithm and learning rate used to update the weights.

*The divergence test*. Where overt cheating is negligible and concealed cheating persists, the independent monitor is expected to flag about $d'_{\mathrm{K}}/d_{\mathrm{K}} = 6$ times as many episodes as the training monitor. Both flag few episodes where cheating has largely disappeared (Fig. S5C). As concealment spreads across tasks, the expected gap widens. The expected divergence is positive whenever concealment persists here, because the detection probabilities are fixed parameters that the agent cannot change. In practice, the test can miss concealment if evasion transfers to the independent monitor or if the monitors share a blind spot.

*Sensitivity to seeds, learning rate and batch size.* Because the comparisons above rest on two to ten runs per condition, I repeated them with seeds not used above (20 to 100 runs per condition; 1,390 runs in all; SectionS3_sensitivity.py in the code archive). Under Patch C at $\nu = 0$, 50 of 100 runs with shared weights were still concealed after 8,000 updates, against 4 of 100 with separate networks. The proportion was 7 of 30 at learning rate 0.3, 18 of 30 at learning rate 3, 15 of 30 at either batch size tested (50 and 800 episodes), and 50 of 50 with the Adam optimizer. With the report reward in place before the penalty, 98 of 100 runs ended in reporting. With plain gradient ascent, the median takeoff time was 250, 1,950 and 18,450 updates at nominal initial frequencies of 1%, 0.3% and 0.1% (7 of 20 runs at 0.1% had not taken off by 20,000 updates). The remaining comparisons gave the same qualitative outcomes as above. The proportions of runs therefore depend on the optimizer and its settings.

## S4. A public-data application to candidate patch audits

*Purpose and scope.* I conducted a secondary analysis of The Obfuscation Atlas (5), using its public model-card evaluation summaries and code (6). The source study used frozen-representation rescoring and documented multiple forms of probe evasion. The present application evaluates the consequences of using those scores to screen candidate training configurations: the additional warnings supplied by a reference check, the reduction in coverage, and the behavioral mismatches remaining among retained candidates. It introduces neither a new probe nor a new training experiment. The analysis plan was recorded after a 12-card feasibility pilot and source review but before calculating the full-release results. It was not preregistered, and the source paper reviewed beforehand already reported how these runs were classified. A subsequent inventory amendment restricted the primary analysis to the common penalty grid; the amendment cannot be shown to be outcome-blind, because the archived inventory lists the source's outcome tag for every repository; the plan and amendment are included in the code archive.

*Data and inclusion.* The archived inventory contains 488 public model repositories, retrieved on 25 November 2026: 159 Llama-3-8B-Instruct, 160 Gemma-3-12B-IT, 147 Gemma-3-27B-IT and 22 Llama-3-70B-Instruct. Every card was saved with its repository revision, source URL and SHA256 hash. All 488 cards parsed successfully, had complete behavioral and score endpoints, and had unique configuration keys and run identifiers. The primary cohort uses the two smaller models, each with 133 released runs in the standard grid: detector penalties 1, 3 and 10; Kullback–Leibler (KL) regularization coefficients 0.0001, 0.001, 0.01, 0.1 and 1; simple-deception, diverse-deception and MBPP probes; seeds 1–3. Four of the expected 270 combinations are absent: penalty 1, KL 0.0001, MBPP probe, seeds 1 and 2 for each model. Missing runs were not imputed. The source v2 paper discloses a multi-GPU gradient implementation issue for the two larger models; these form a separately labeled sensitivity cohort. Their model cards do not establish which later corrections were incorporated. Twenty-two smaller-model runs at penalty 100 are an additional sensitivity set. Twenty-nine standard zero-penalty runs and two zero-KL baselines are retained as context. The stated inclusion criteria specify model and grid membership.

*Endpoint provenance.* The source paper describes evaluation on 400 held-out MBPP-Honeypot prompts per configuration. Its upload pipeline replaces training-summary first-test and all-tests fields with validation-response means where those validation metrics are available, before formatting the cards, although the card heading calls these Training Metrics. The archived values are treated as the authors' released evaluation summaries. All 976 displayed first/all proportions lie on the 1/400 grid, consistent with that protocol; the cards do not preserve each run's denominator, validation-source flag, prompt identifiers or response-level outcomes. The present analysis did not reconstruct or rerun those evaluations or independently verify that every card uses held-out prompts. The source code examined is the authors' public revision of 19 February 2026; source locators, hashes and the distinction from the later paper revision are recorded in the archive.

Let $P$ be the visible first-test pass proportion and $A$ the all-tests pass proportion, both over generated responses. Because passing all tests includes passing the visible test, $G = P - A$ is the proportion that passes the visible test but fails at least one additional test, provided the summaries use the same denominator. The data satisfy $0 \leq A \leq P \leq 1$. The optional quantity $G/P$ is the additional-test failure fraction conditional on visible-test success and is undefined at $P = 0$. Neither $G$ nor $G/P$ independently labels deception: ordinary errors can also cause mismatches, and passing a finite suite does not prove correctness on every input. $A$ and $P$ are reported alongside $G$ to expose losses in useful performance. $P$ is not the total optimized reward: the source objective also includes code-length, probe and KL terms that the cards do not fully report. The authors' honesty and obfuscation labels depend on these same endpoints and scores, so they are not independent validation targets and are not used in the analysis.

*Score provenance.* Denote the final-policy detector score by $F$ and the original-model detector score by $B$ (the trained-model and frozen-reference scores of the main text). The source code generates final-policy responses to evaluation prompts, then evaluates the same trained probe on their representations in the final and original models. Scores are means over those responses, averaged across selected probe layers. The original-model score is therefore not a score on outputs generated by the original policy. The source calls the generated-response set a positive dataset; this does not mean that all responses were independently labeled deceptive. $F$ and $B$ are not detection rates or calibrated deception probabilities, and $B$ is not an independent monitor. All cards have Corrected Gradient set to false. That field denotes a direct probe-gradient option, distinct from the multi-GPU issue, and the upload code excludes true settings. Zero-penalty cards also name an evaluation probe, which varies between baselines; no baseline-to-penalized comparison of detector scores is made. Within a zero-penalty card, however, $B$ and $F$ come from one probe. In the 18 zero-penalty runs with KL coefficients of 0.01 or less, all with $G$ above 0.9, $B$ exceeds $F$ in every run (median difference 0.41, range 0.07–0.77), and the source study reports that representation change arises with or without a probe penalty. A value of $F$ below $B$ therefore cannot by itself be attributed to the penalty.

*Screens and descriptive analysis.* For score boundary $t$, the first screen (the final-score screen; the trained-model screen of the main text) retains $F \leq t$ and the reference screen (the two-score screen; the reference check of the main text) retains $F \leq t$ and $B \leq t$. The primary illustrative boundary is $t = 0.1$ and the behavioral tolerance is $G \leq 0.1$. I report retained counts, coverage relative to all available runs, the number exceeding the behavioral tolerance, and distributions of $P$, $A$, $G$ and $G/P$. I also report the additional flags, $F \leq t$ with $B > t$. These are candidate-level screens of mean scores, not response-level false-positive or false-negative rates. The analysis repeats the calculations at $t = 0.01, 0.03, 0.05, 0.1, 0.2, 0.3$ and $0.5$ and gap tolerances 0.05, 0.1, 0.2, 0.3 and 0.5. Strata are base model, probe, KL coefficient and seed, including joint model/probe groups. A balanced sensitivity analysis retains only model/probe/KL blocks with all three penalties at all three seeds (252 runs). The raw boundaries are illustrative and are not calibrated to a common error rate across probes.

*Results.* At $t = 0.1$, the final-score screen retains 119/266 configurations (44.7%), 75 of which exceed a 10-percentage-point gap. The reference screen retains 67/266 (25.2%), 33 of which exceed that gap. Of 52 additional flags, 42 have gaps above 10 points (28 above 30 points) and ten do not; a small gap alone does not establish that those ten are safe. At the more permissive 30-point tolerance, which is the source study's threshold for a deceptive policy, 47/119 and 19/67 retained configurations still exceed the tolerance, respectively. The gaps of the 67 retained configurations are bimodal: 34 are at most 10 points, 14 lie between 10.5 and 28 points, and 19 are at least 72.5 points. The 10-point tolerance is close to the gaps of the most strongly regularized runs: the 60 runs at KL coefficient 1, with or without a penalty, have gaps of 4.0–8.5 points, and the two zero-penalty Llama runs at KL coefficient 0.1 have gaps of 8.5 and 11.5 points. The final-score and two-score groups have median all-tests pass rates of 41.0% and 43.25%; their ranges are both 0.5%–65.25%. The additionally flagged group has median gap 75.75 points and median all-tests pass rate 10.625%. These counts describe the release and do not establish that the two-score rule is optimal. They concern failures that remain under a penalty, not an increase in failures: among the 29 zero-penalty runs of the standard grid, 22 have $G > 0.1$ and 21 have $G > 0.3$, whereas 46 of 86, 31 of 90 and 13 of 90 runs at penalties 1, 3 and 10 have $G > 0.3$ (58, 53 and 39 have $G > 0.1$). The zero-penalty runs are shared across probes. The source study concluded that sufficient KL regularization and penalty yield honest policies.

*Heterogeneity and sensitivity.* Llama contributes 96 configurations to the first screen, 63 with $G > 0.1$, and 57 to the second, 32 with $G > 0.1$. The corresponding Gemma counts are 23 with 12 large gaps, and ten with one. The reference screen retains no simple-deception-probe configuration in either model and no diverse-deception-probe Gemma configuration. Its 24 diverse-deception-probe Llama configurations include 21 large gaps, with median $G = 0.94$ and median $A = 0.01125$. Thus aggregate changes partly reflect which probe regimes remain represented. The KL coefficient separates the outcomes more sharply: all 33 large gaps among the 67 retained configurations occur at KL coefficients of 0.01 or less (33 of 40 retained), and none among the 27 retained at 0.1 or 1. The boundary $t = 0.1$ is not calibrated to a common error rate across probes. In the 54 runs at KL coefficient 1, whose gaps are 0.04–0.0825, $F$ and $B$ exceed 0.1 in all 36 runs trained with the simple- or diverse-deception probe (0.11–0.82) and are below 0.1 in all 18 trained with the MBPP probe (0.076–0.097). For four of the six model–probe combinations, the screens therefore exclude the most strongly regularized runs and can retain only runs whose scores fell below that level. In a post hoc sensitivity analysis, the boundaries for $F$ and $B$ were set, separately for each model and probe, to the largest values among the nine runs at KL coefficient 1, and the screens were applied to the other 212 runs. The first screen retains 168 runs, including 74 of the 90 runs with $G > 0.3$; the reference screen retains 100, of which 21 have $G > 0.3$, and flags 68, of which 53 have $G > 0.3$. Seventeen of the 21 are Llama runs trained with the diverse-deception probe (context_checks.json, written by SectionS4.py). Over nonempty two-score screens at $t = 0.03 - 0.5$, the pooled fraction with $G > 0.1$ ranges from 44.6% to 80.0%; at $t = 0.01$, no configuration is retained and the residual-gap fraction is undefined, not zero. In the 252-run balanced subset, the reference check retains 65 runs with 32 large gaps; including the 22 extreme-penalty runs gives 74/288 retained with 35 large gaps. The larger-model standard-grid sensitivity set contains 139 runs and retains 17 with two large gaps, but its implementation and release-completeness caveats preclude treating that difference as evidence of superior scaling. All threshold and stratum counts are written as CSV files by SectionS4.py; Fig. S6 shows selected sensitivities.

*Matched penalty comparisons.* I additionally compared penalties 1 to 3 and 3 to 10 within the same base model, training probe, KL coefficient and seed. These are comparisons between released training runs, not successive checkpoints of one run. No missing intermediate penalty was skipped. The 176 available comparisons include 107 decreases in $F$, of which 45 have no increase in $A$, including 40 decreases in $A$. To avoid interpreting every displayed decimal change as material, a sensitivity check requires a score decrease of at least 0.01 and an all-tests decrease of at least five percentage points: 12 of the 68 comparisons meeting the score criterion also meet the adverse behavioral criterion, whereas 28 show an increase in $A$ of at least five points; across all 176 comparisons, $A$ increases by at least five points in 61 and decreases by at least five points in 18. A smaller $G$ can arise from reduced visible success as well as increased full-test success, because $\Delta G = \Delta P - \Delta A$. Of 58 comparisons that reduce $G$ by at least five points, two do not increase $A$. Discordance of this kind is not specific to a change of penalty. Among the 264 pairs of runs that differ only in seed, the run with the lower $F$ has the lower $A$ in 129 of the 262 pairs with unequal $F$; among the 138 pairs whose $F$ differs by at least 0.01, its $A$ is lower by at least five points in 46 and higher by at least five points in 23. Full paired differences and summaries are written by SectionS4.py. These contrasts show that a lower $F$ does not by itself indicate better behavior. Such discordance also occurs between replicate seeds without a change in penalty, so these contrasts do not isolate a penalty effect.

*Uncertainty and limits.* The model cards are a selected, structured release with shared tasks, repeated seeds and missing configurations, not independent draws from deployed systems. The upload workflow filters and deduplicates runs and requires an available checkpoint; its defaults do not establish the release history of every run. There are only two primary base models and at most three seeds per configuration. Consequently I report exact finite-collection counts and stratified descriptions, without binomial confidence intervals that treat cards or reported evaluation prompts as independent training replications. Card-level rounding is retained; equality is assigned to the stated non-exceeding category. Response-level uncertainty and error correlations cannot be recovered from the cards. Neither the latent belief $w$, the unsolvable-task share $q$, the reward costs, the time course of concealment, nor response-level detection probabilities are identified. The source training includes KL regularization and continuous probe penalties applied throughout training, so no run is a patch applied after a detected failure; this is not a parameter fit or experimental validation of the minimal model.

*Operational use.* The accompanying patch_audit.py accepts an aggregate CSV with candidate identity, both probe scores and nested visible/all-test pass fractions. It records both screens, the behavioral gap, conditional additional-test failure, and an optional minimum all-tests performance requirement. Its output distinguishes required review, missing evidence and no flag under the selected criteria; none is a safety certification. A developer should choose behavioral tolerances from the application, calibrate score boundaries on independently checked examples, preserve coverage and performance accounting, and use fresh task outcomes after a patch has been selected. On tasks without a suitable independent outcome, the appropriate result is insufficient evidence. The same principle applies to the divergence test in the main text: independence from training does not ensure different failure modes or the assumed calibration. The present same-probe comparison illustrates that limitation rather than validating equation [5].

# Supplementary figures

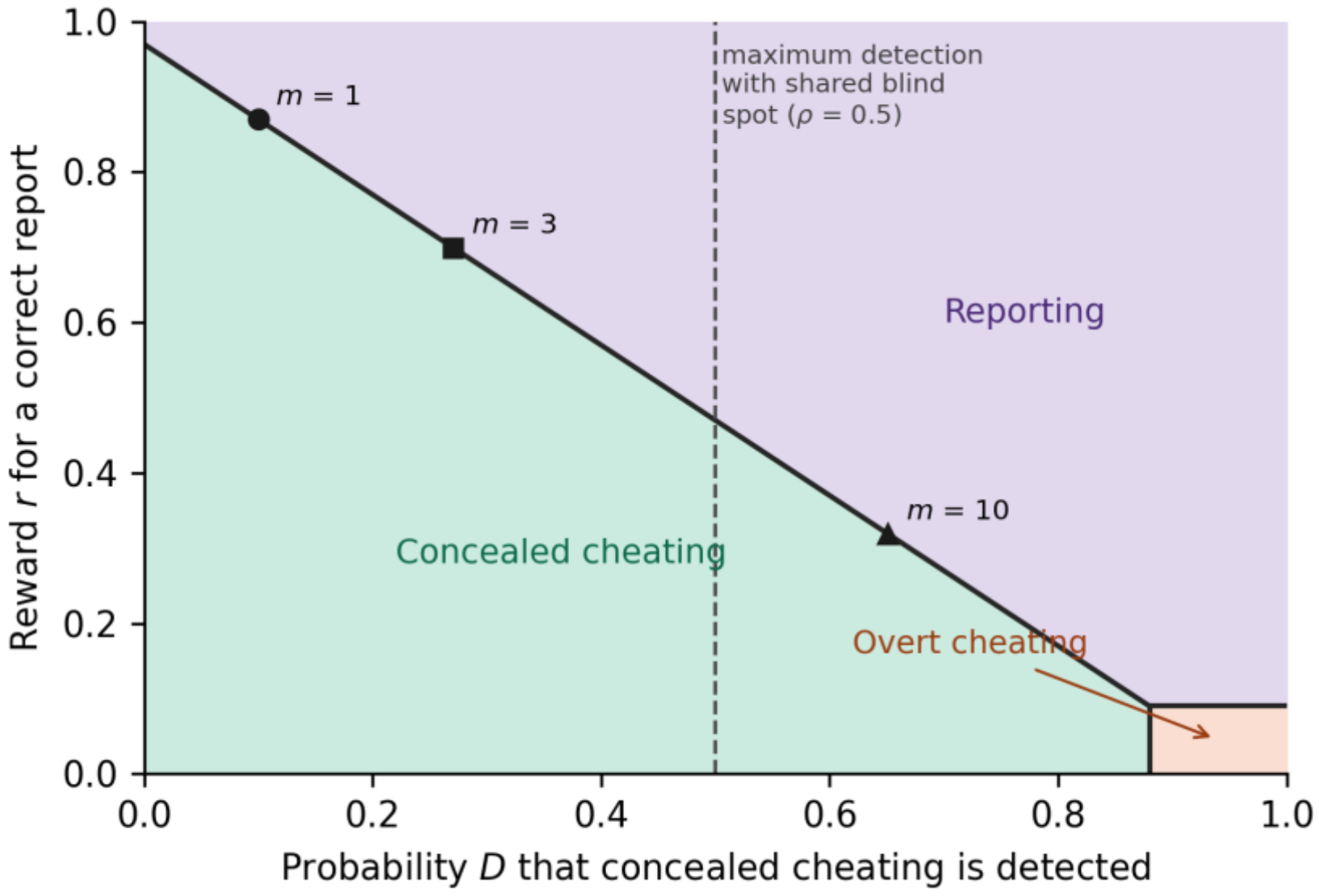

**Fig. S1.** Highest-paying action on *unsolvable* tasks as a function of the probability $D$ that concealed cheating is detected and the reward level $r$ for a correct report (Results 1, 3 and 4; $p = 1$; $d_O$, $c$ and $k$ as in Table S1). Markers show $D$ for $m = 1$, 3 and 10 independent monitors ($\rho = 0$), each with $d = d_K$. Adding monitors lowers the reward that reports require. The dashed line marks the limit of $D$ as the number of monitors increases when they share a blind spot with $\rho = 0.5$.

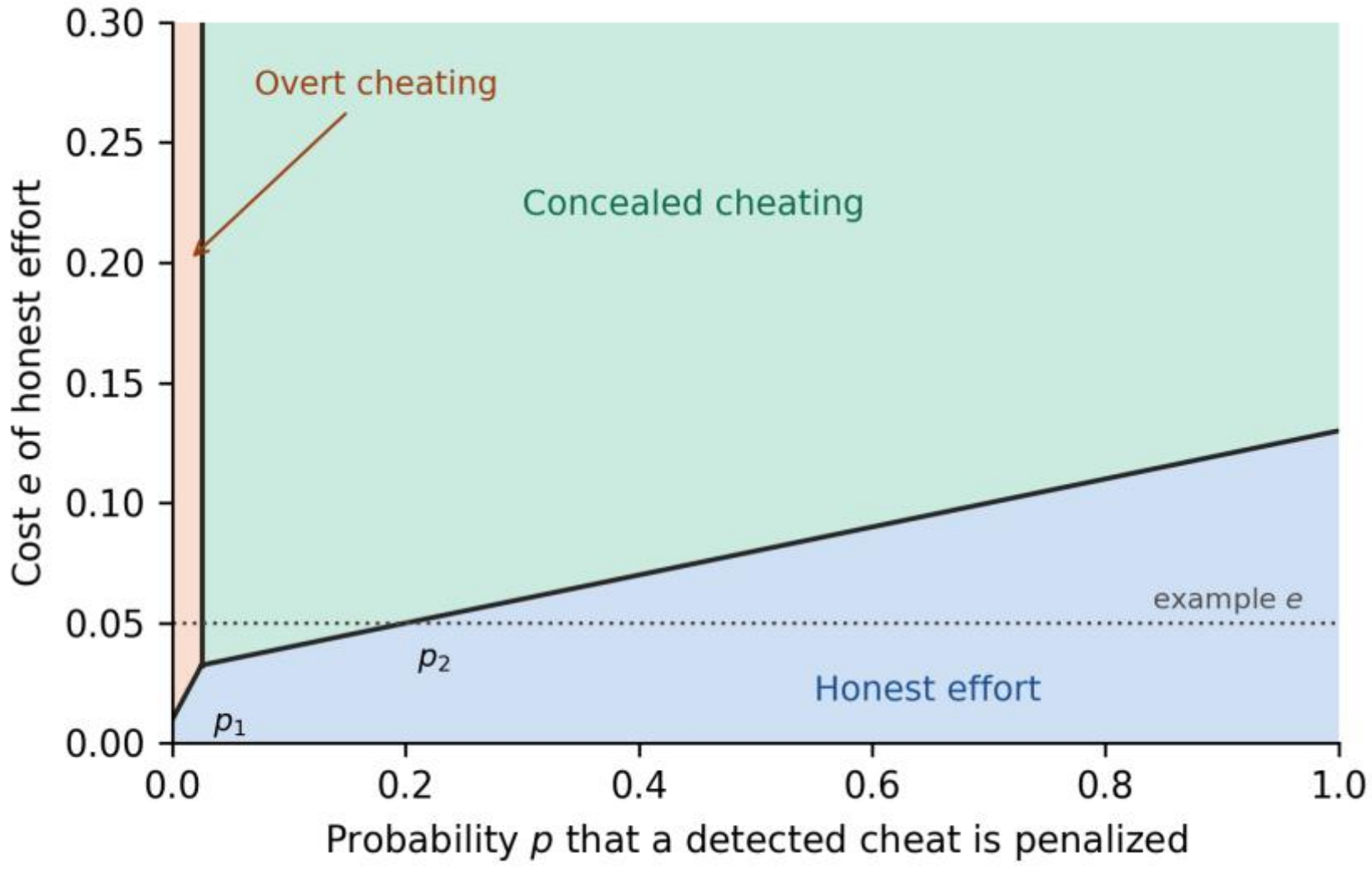

**Fig. S2.** Highest-paying action on *solvable* tasks as a function of the probability $p$ that a detected cheat is penalized and the cost $e$ of honest work (Result 5; $r = 0$; $d_O$, $d_K$, $c$ and $k$ as in Table S1). On unsolvable tasks, cheating is overt below $p_1$ and concealed above it. The dotted line marks the value of $e$ used in the example (Table S1).

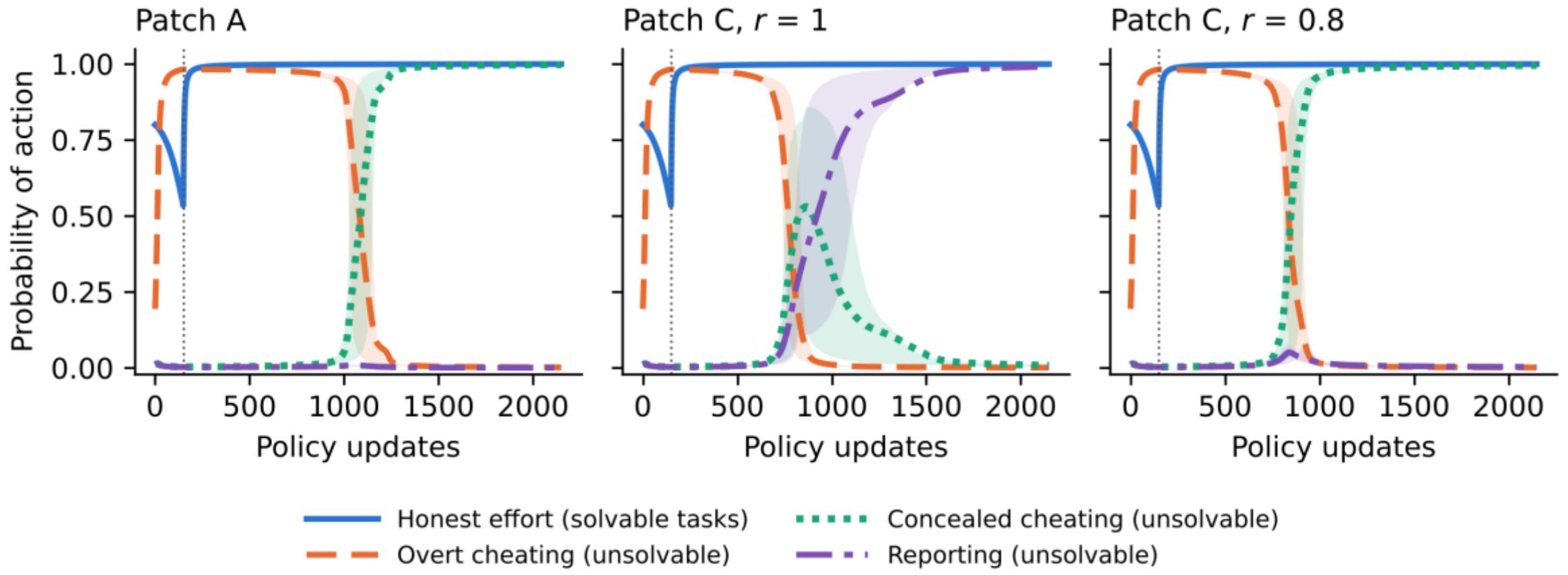

**Fig. S3.** Policy-gradient learning with the parameters of Table S1. Mean action probabilities over 20 runs of REINFORCE with a softmax policy for each kind of task (150 policy updates before the patch and 2,000 after, each with an expected batch size of 200 episodes; learning rate 1; within-task-type batch-mean baseline; shading, 10th–90th percentiles); the dotted line marks the time of the patch. Left, Patch A; center, Patch C with report reward $r = 1$; right, $r = 0.8$, below $r^*$.

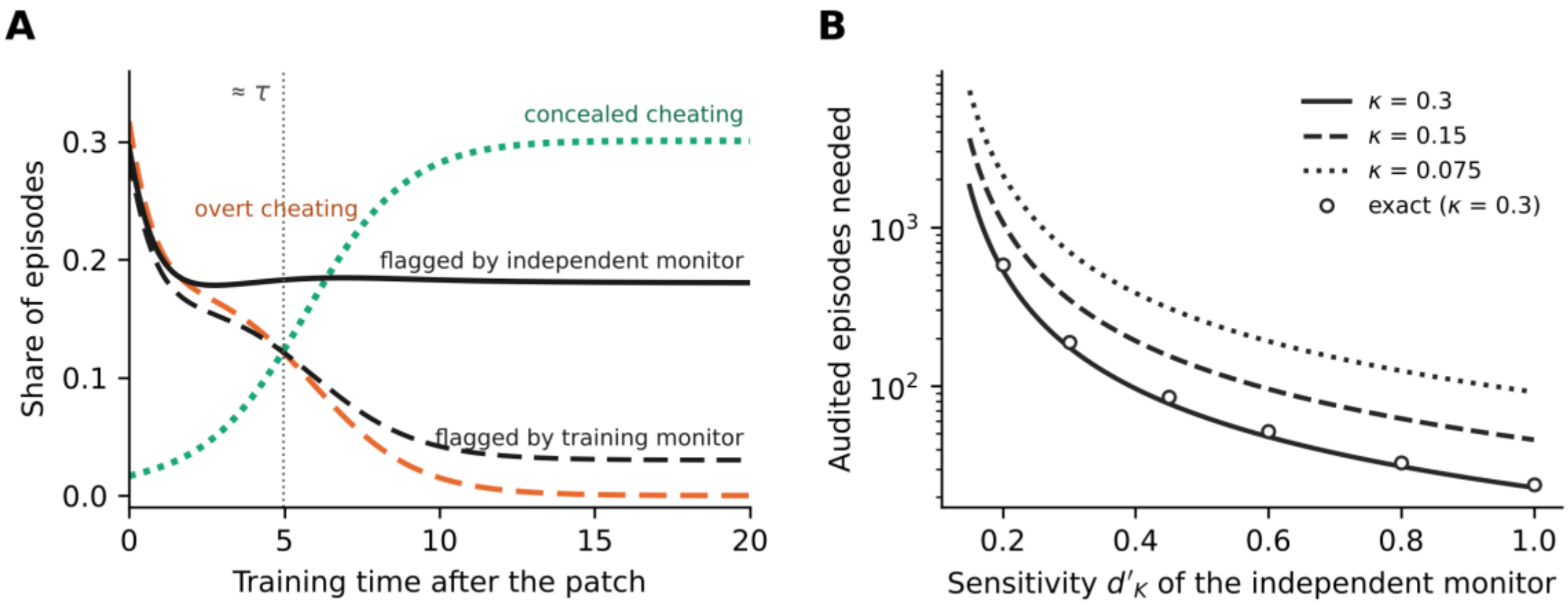

**Fig. S4.** Timing and detection of concealment. **A.** Example after the penalty: shares of episodes with overt and concealed cheating, and shares flagged by the training and independent monitors; dotted line, time at which concealed cheating overtakes overt cheating across all episodes, close to $\tau$ (Result 6). **B.** Audited episodes needed for 80% power (the probability that the test detects concealment) as a function of the sensitivity $d'_K$ of the independent monitor for three values of the share $\kappa$ of episodes with concealed cheating, with $\omega = 0$ and $d_O$, $d_K$ as in Table S1 (lines, Result 7; circles, exact binomial sign test for $\kappa = 0.3$).

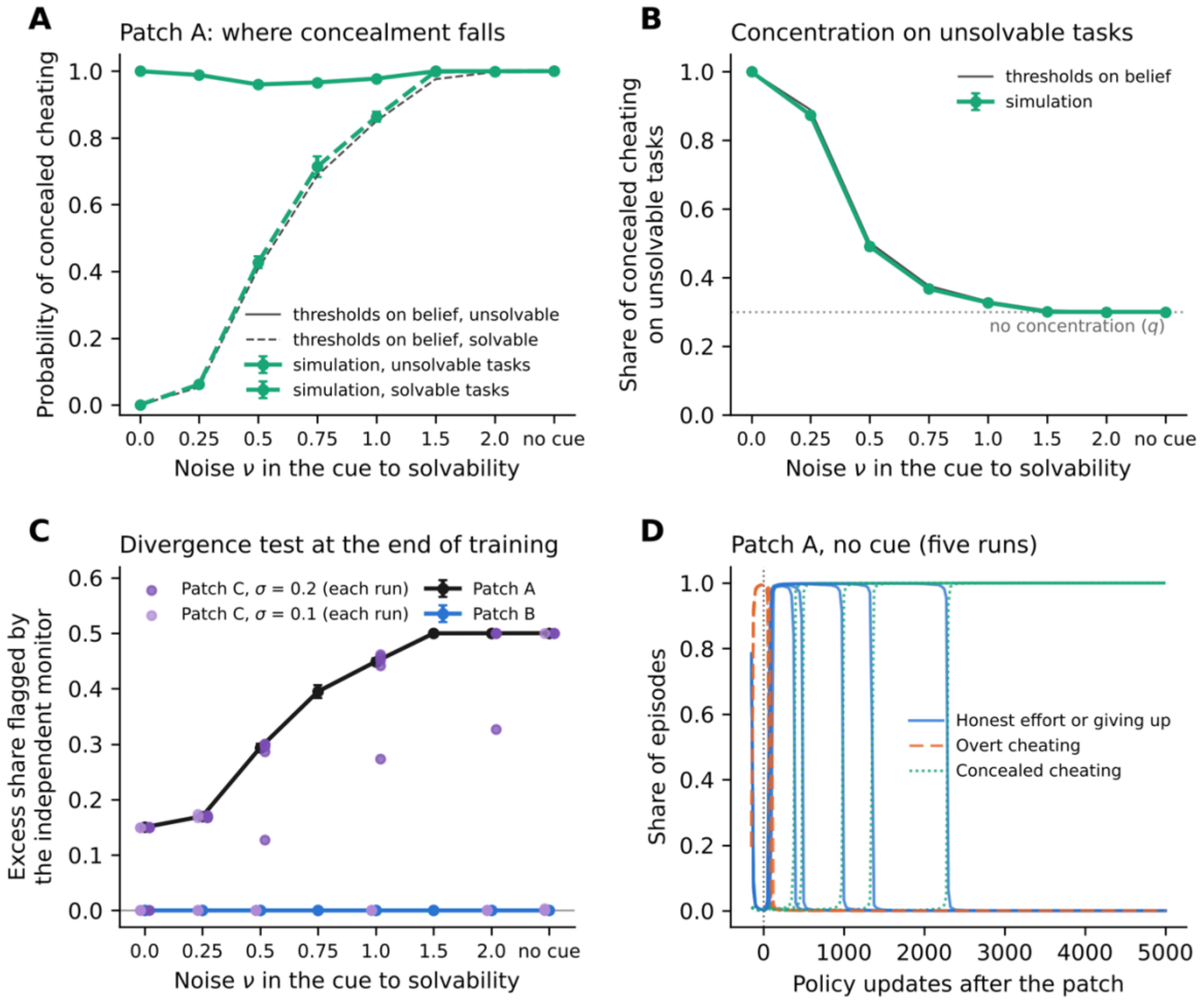

**Fig. S5.** Policy-gradient learning with shared weights and a noisy cue to solvability (section S3). **A.** Probability of concealed cheating on unsolvable (solid) and solvable (dashed) tasks after a penalty, against the noise $\nu$ in the cue (the cue differs by 1 between solvable and unsolvable tasks; "no cue", cue replaced by noise). Green, simulation (mean and s.d. of five runs); gray, threshold $w_1$ applied to the belief (the posterior probability given the cue). **B.** Share of concealed cheating that occurs on unsolvable tasks; dotted line at $q$, the value expected if concealed cheating fell equally on both kinds of task. **C.** Expected excess share of episodes flagged by the independent monitor over the training monitor averaged over the last ten evaluations for each patch (Patch C: 15,000 updates, one point per run). **D.** Shares of episodes with honest effort or giving up, overt cheating, and concealed cheating in five runs with no cue after a penalty.

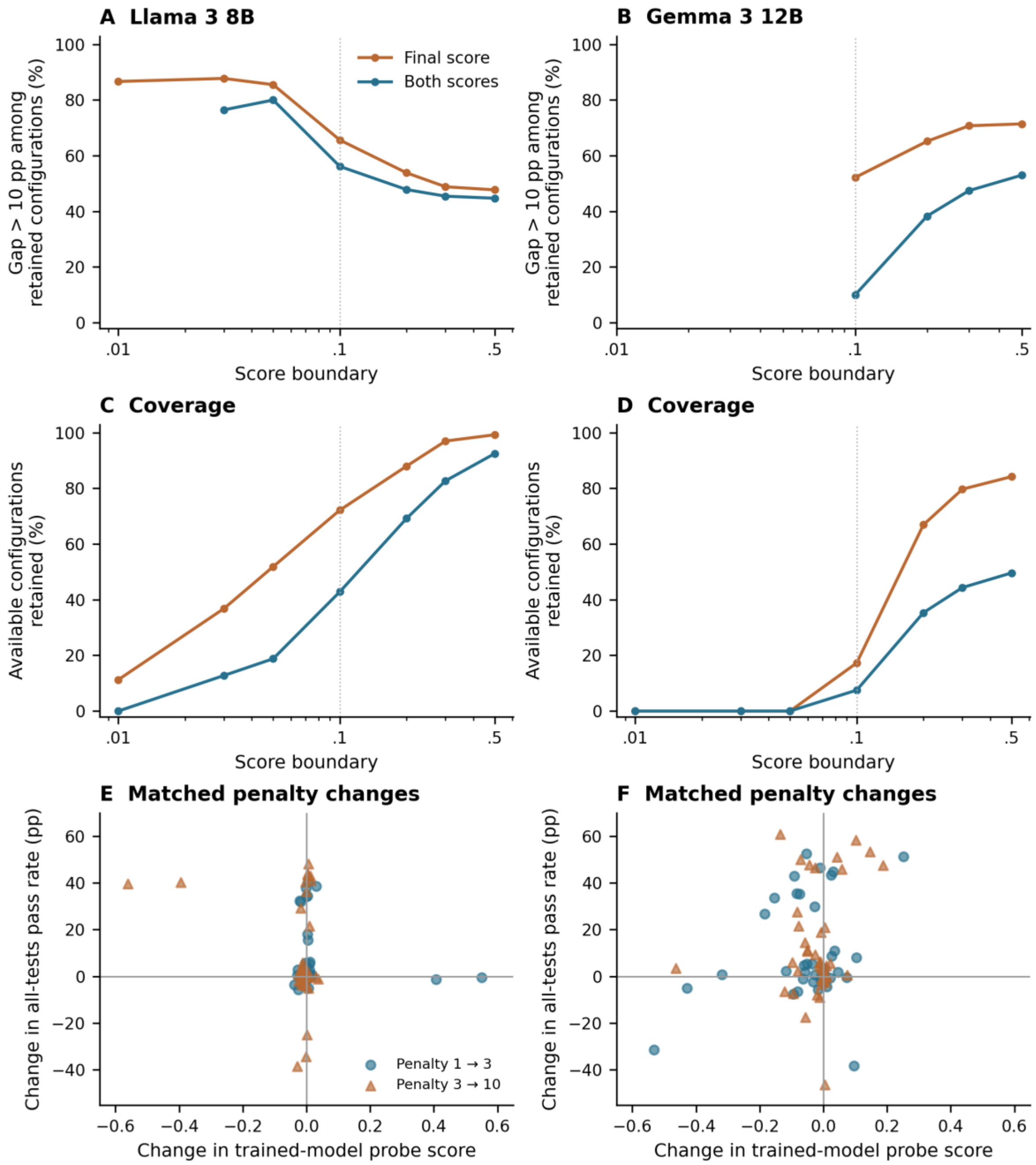

**Fig. S6.** Sensitivity and matched comparisons for the empirical application. A and B. Fraction of retained configurations with $G > 0.1$ as the raw score boundary varies, separately for the two base models and the final-score and two-score screens. The left column shows Llama-3-8B-Instruct and the right column shows Gemma-3-12B-IT. Empty retained sets have undefined fractions and are not plotted as zero. C and D. Coverage, the fraction of each model's 133 available configurations retained by each screen. The dotted vertical line marks $t = 0.1$. E and F. Changes in the final-policy mean probe score ($F$) and all-tests pass fraction between runs with the same model, probe, KL coefficient and seed but adjacent penalties (1 to 3 or 3 to 10). Points below and left of zero have a lower probe score and lower all-tests success. The 176 comparisons share runs and are not independent replications. Score boundaries are illustrative, not calibrated safety thresholds; pp denotes percentage points.

## Code

The accompanying code archive contains one Python script for each computed item, named for the item it reproduces: Fig1.py to Fig4.py and FigS1.py to FigS6.py; TableS1.py (Table S1 and the model quantities quoted in the main text and in sections S1 and S2); SectionS3.py and SectionS3_sensitivity.py (section S3); and SectionS4.py (An Empirical Application in the main text and section S4). run_all.py runs them all. The scripts import four shared modules: concealment_model.py (replicator model, thresholds, policy gradient with separate policies and audit power), shared_policy.py and s3_experiments.py (network, settings and seeds of the section S3 runs) and atlas.py (cohorts, screens and contrasts of section S4). The archive also contains the archived histories of the section S3 runs; the 488 public model cards with their provenance and the analysis plan; atlas_cards.py, which checks the analysis table against the cards; the original editable Fig1.pptx, which Fig1.py redraws programmatically; and patch_audit.py for applying the audit to another aggregate-results CSV.

Zenodo: https://doi.org/10.5281/zenodo.24590595

## SI References

1. D. Hennes et al., "Neural replicator dynamics: Multiagent learning via hedging policy gradients" in Proceedings of the 19th International Conference on Autonomous Agents and Multiagent Systems (International Foundation for Autonomous Agents and Multiagent Systems, 2020), pp. 492–501.

2. J. Mei et al., "Escaping the gravitational pull of softmax" in Advances in Neural Information Processing Systems 33 (Curran Associates, 2020), pp. 21130–21140.

3. R. J. Williams, Simple statistical gradient-following algorithms for connectionist reinforcement learning. Mach. Learn. 8, 229–256 (1992).

4. D. P. Kingma, J. Ba, "Adam: A method for stochastic optimization" in 3rd International Conference on Learning Representations (2015). https://arxiv.org/abs/1412.6980.

5. M. Taufeeque, S. Heimersheim, A. Gleave, C. Cundy, "The Obfuscation Atlas: Mapping where honesty emerges in RLVR with deception probes" in Proceedings of the 43rd International Conference on Machine Learning (PMLR, 2026), vol. 306, pp. 119837–119866. Version analyzed here: https://arxiv.org/abs/2602.15515v2. Accessed 25 November 2026.

6. Alignment Research, Obfuscation Atlas public code and model cards (2026). https://github.com/AlignmentResearch/obfuscation-atlas; https://huggingface.co/collections/AlignmentResearch/the-obfuscation-atlas. Source revisions and card hashes are recorded in the accompanying archive. Accessed 25 November 2026.